\newcommand{\etal}{{\em{et al. }}}       % et al.
\newcommand{\eg}{{\em e.g.}}           % e.g.
\newcommand{\ie}{{\em i.e.}}  
\newcommand{\etc}{{\em etc}}         % etc.

\newcommand{\BT}[1]{\textit{\textbf{#1}}}
\documentclass[sigconf]{acmart}
\usepackage{colortbl} % 确保已包含colortbl包
\usepackage{bigstrut, multirow, rotating}
\usepackage{pifont}
\usepackage{makecell} 
\usepackage{enumitem}
\AtBeginDocument{%
  }

\setcopyright{acmlicensed}
\copyrightyear{2025}
\acmYear{2025}
\setcopyright{acmlicensed}\acmConference[MM '25]{Proceedings of the 33rd ACM International Conference on Multimedia}{October 27--31, 2025}{Dublin, Ireland}
\acmBooktitle{Proceedings of the 33rd ACM International Conference on Multimedia (MM '25), October 27--31, 2025, Dublin, Ireland}
\acmDOI{10.1145/3746027.3754973}
\acmISBN{979-8-4007-2035-2/2025/10}

\begin{document}

%%
%% The "title" command has an optional parameter,
%% allowing the author to define a "short title" to be used in page headers.
\title{Towards Perfection: Building Inter-component Mutual Correction for Retinex-based Low-light Image Enhancement}
\settopmatter{authorsperrow=4}
%%
%% The "author" command and its associated commands are used to define
%% the authors and their affiliations.
%% Of note is the shared affiliation of the first two authors, and the
%% "authornote" and "authornotemark" commands
%% used to denote shared contribution to the research.
\author{Luyang Cao}
\affiliation{%
  \institution{Nanjing University}
  \city{Nanjing}
  \country{China}}
\email{caoluyang@smail.nju.edu.cn}

\author{Han Xu}
\affiliation{%
  \institution{Southeast Univeristy}
  \city{Nanjing}
  \country{China}}
\email{xu\_han@seu.edu.cn}

\author{Jian Zhang}
\affiliation{%
  \institution{Nanjing University}
  \city{Nanjing}
  \country{China}}
\email{zhang.jian@nju.edu.cn}

\author{Lei Qi}
\affiliation{%
  \institution{Southeast Univeristy}
  \city{Nanjing}
  \country{China}}
\email{qilei@seu.edu.cn}

\author{Jiayi Ma}
\affiliation{%
  \institution{Wuhan University}
  \city{Wuhan}
  \country{China}}
\email{jiayima@whu.edu.cn}

\author{Yinghuan Shi}
\authornote{Corresponding Author.}
\affiliation{%
  \institution{Nanjing University}
  \city{Nanjing}
  \country{China}}
\email{syh@nju.edu.cn}

\author{Yang Gao}
\affiliation{%
  \institution{Nanjing University}
  \city{Nanjing}
  \country{China}}
\email{gaoy@nju.edu.cn}

%%
%% By default, the full list of authors will be used in the page
%% headers. Often, this list is too long, and will overlap
%% other information printed in the page headers. This command allows
%% the author to define a more concise list
%% of authors' names for this purpose.
\renewcommand{\shortauthors}{Luyang Cao et al.}

%%
%% The abstract is a short summary of the work to be presented in the
%% article.
\begin{abstract}
In low-light image enhancement, Retinex-based deep learning methods have garnered significant attention due to their exceptional
interpretability. These methods decompose images into mutually
independent illumination and reflectance components, allows each
component to be enhanced separately. In fact, achieving perfect decomposition of illumination and reflectance components proves to be quite challenging, with some residuals still existing after decomposition. In this paper, we formally name these residuals as inter-component residuals (ICR), which has been largely underestimated by
previous methods. In our investigation, ICR not only affects the accuracy of the decomposition but also causes enhanced components to deviate from the ideal outcome, ultimately reducing the final synthesized image quality.
To address this issue, we propose a novel Inter-correction Retinex model (IRetinex) to alleviate ICR during the
decomposition and enhancement stage. In the decomposition stage, we leverage inter-component residual reduction module  to reduce the feature similarity between illumination and reflectance components. In the enhancement stage, we utilize the feature similarity between the two components to detect and mitigate the impact of ICR within each enhancement unit.
Extensive experiments on three low-light benchmark datasets demonstrated that by reducing ICR, our method outperforms state-of-the-art approaches both qualitatively and quantitatively. Our code is available at:  \color{magenta}{https://github.com/caoluyang0830/IRetinex.git}.
\vspace{-0.2cm}
\end{abstract}

%%
%% The code below is generated by the tool at http://dl.acm.org/ccs.cfm.
%% Please copy and paste the code instead of the example below.
%%


%%
%% Keywords. The author(s) should pick words that accurately describe
%% the work being presented. Separate the keywords with commas.
\vspace{-0.1cm}
\vspace{-0.1cm}
%% A "teaser" image appears between the author and affiliation
%% information and the body of the document, and typically spans the
%% page.
% \begin{teaserfigure}
%   \includegraphics[width=\textwidth]{sampleteaser}
%   \caption{Seattle Mariners at Spring Training, 2010.}
%   \Description{Enjoying the baseball game from the third-base
%   seats. Ichiro Suzuki preparing to bat.}
%   \label{fig:teaser}
% \end{teaserfigure}

% \received{20 February 2007}
% \received[revised]{12 March 2009}
% \received[accepted]{5 June 2009}

%%
%% This command processes the author and affiliation and title
%% information and builds the first part of the formatted document.
\maketitle

\vspace{-0.1cm}
\section{Introduction}\label{Introduction}
In low-light condition, image quality degrades significantly, manifesting as low contrast, low visibility, dense noise, color distortions, \etc~\cite{Wang_IDP_2024}. In real applications, low-quality images  not only impair human visual perception but also hinder computer vision systems in performing high-level image processing tasks, \eg, detection and recognition~\cite{fengLearnabilityEnhancementLowLight2024}. To address this challenge of multimodal perception optimization, low-light image enhancement (LLIE) has been proposed to correct color distortions and recover buried details in low-light conditions~\cite{liLowlightImageVideo2021a,wangExposureDiffusionLearningExpose, xucretinex2024}, thereby preserving the natural perception of human visual modality and simultaneously satisfying the demand of computer vision modality for feature representation.

\begin{figure}[!t]

\centering
\includegraphics[width=0.96\columnwidth]{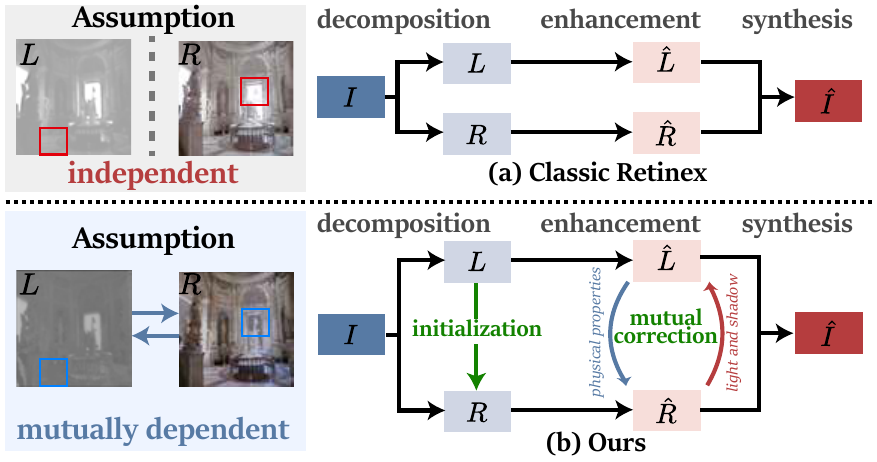}
\vspace{-0.4cm}
\caption{Comparison between the classic Retinex methods and our method.  
(a) The classic Retinex methods methods independently process illumination/reflectance, generating inter-component artifacts (red boxes).  (b) Our method employs mutual correction during decomposition and enhancement, mitigating residuals and achieving satisfactory results.}
\vspace{-0.4cm}
\label{fig:firstframework}
\end{figure}

With the development of deep learning, learning-based methods have become the mainstream approach for LLIE~\cite{fuYouNotNeed2023,yangSparseGradientRegularized2021,caiRetinexformerOnestageRetinexbased}. Among these approaches, the methods based on Retinex theory~\cite{wangProgressiveRetinexMutually2019,caiRetinexformerOnestageRetinexbased,liuRetinexinspiredUnrollingCooperative2021} have garnered significant attention due to its promising results.
% since it decouples different degradations (\eg, ??) from a single image for easier enhancement.
Mathematically, Retinex theory decomposes the image \( I \) into illumination \( L \) and reflectance \( R \) components:
\vspace{-0.1cm}
\begin{equation}
    I = L \odot R,
    \label{eq:retinex}
\end{equation}

\vspace{-0.1cm}
\noindent where \( \odot \) represents element-wise multiplication.
As shown in Fig.~\ref{fig:firstframework}, based on Retinex theory, low-light image enhancements contain three stages: 1) a \textit{decomposition} stage to decouple the low-light image $I$ into illumination component $L$  (\ie, global light condition) and reflectance component $R$ (\ie, physical properties including color and texture, \etc), 2) an \textit{enhancement} stage to adjust illumination $\hat{L}$ and reflectance $\hat{R}$ components separately, and 3) a \textit{synthesis} stage to combine two components for the normal-light image $\hat{I}$.
Taking advantage of the decomposition, Retinex-based methods effectively adapt to various lighting conditions and handle reflectance degradation, thereby achieving visually appealing results~\cite{fuYouNotNeed2023}.

\begin{figure}[!t]
\vspace{-0.3cm}
\centering
\includegraphics[width=0.96\columnwidth]{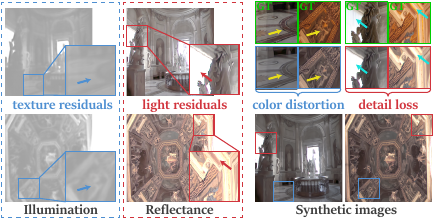}
\vspace{-0.3cm}
\caption{Method motivation. {\BT{Imperfect decomposition leads to residuals remaining in the separated components.}} These residuals subsequently cause color distortion and detail loss in the synthesized images. GT: ground truth.}
\label{fig:firstfig}
\vspace{-0.4cm}
\end{figure}

Despite their success, we found that existing methods are based on a fundamental yet idealized assumption: \BT{illumination and reflectance are mutually independent}~\cite{zhangBrighteningLowlightImages2021a,liuRetinexinspiredUnrollingCooperative2021}. Specifically, the reflectance component is required to be devoid of light or shadow (\ie, illumination information), while the illumination component should be free from physical properties (\ie, reflectance information). This assumption implies that, after decomposition, illumination and reflectance components should be perfectly separated, thus allowing their independent enhancement. However, in real cases,  \textit{Retinex decomposition is a highly ill-posed problem and there is no available ground truth for the illumination and reflectance components~\cite{fuYouNotNeed2023}, which makes perfect decomposition challenging to achieve.}
% . {Previous methods attempt to reduce the decomposition difficulty by introducing additional manually designed priors~\cite{zhangKindlingDarknessPractical2019, weiDeepRetinexDecomposition2018, renLR3MRobustLowlight2020,Wu_URetinex_plus}. However, \textit{these specifically designed priors inevitably introduce residuals that differ from physical realities},

To further illustrate this issue, we visualize the decomposition results of SOTA Retinex-based learning methods~\cite{Wu_URetinex_plus}. As shown in the left panel of Fig.~\ref{fig:firstfig}, evidently, texture details (part of reflectance) bleed into the illumination component, and light and shadows (part of illumination) remain in the reflectance component (marked by the blue and red boxes in Fig.~\ref{fig:firstfig}). During the enhancement stage, due to the existence of these residuals, the enhancement components inevitably deviate from the ideal outcome, ultimately degrading the quality of the final synthesized image, manifesting as color distortion and detail loss (shown in the right panel of Fig.~\ref{fig:firstfig}). {\BT{In this paper, we investigate this issue that has been largely underestimated by previous methods, providing the first formal definition of these residuals as inter-component residuals (ICR).}}
% impact the accuracy of illumination and reflectance enhancement

Based on above discussion, within the Retinex-based LLIE pipeline, reducing ICRs during the decomposition stage and mitigating their impact during the enhancement stage are crucial.
Fortunately, as illustrated in the left panel of Fig.~\ref{fig:firstfig}, we discovered that ICRs in the illumination component exhibit feature similarities to those in the corresponding reflectance component, and vice versa. This observation naturally prompts us to consider: {\BT{Could we leverage these feature similarities to identify and mitigate ICRs?} }

Intuitively, during the decomposition stage, the low feature similarity between the illumination and reflectance components indicates that each component retains its inherent distinctive features, with no significant residuals. In the enhancement stage, the similar features actually reveal different component missed features. By supplementing these features into corresponding components, we could further drive the illumination and reflectance components closer to their ideal outcomes. Therefore, reducing feature similarity is beneficial to achieving perfect decomposition. Moreover, the similar features could facilitates mutual correction between the two components, thereby mitigating  the ICR impact during the enhancement stage.

Based on the above analysis, we propose a novel inter-correction Retinex model (IRetinex) that systematically addresses ICR issues across both decomposition and enhancement stages. During the decomposition stage, we propose an innovative inter-component residual reduction module designed to minimize feature similarity between illumination and reflectance components, reducing initial decomposed ICR. During enhancement, we propose a novel module for residual mitigation and component enhancement. By leveraging feature similarity, this module identifies and mitigates ICRs across components and recovering buried details in low-light conditions. Additionally, we propose a Retinex-based multi-scale consistency loss that provides targeted supervision for the ICR identification process. Our contributions are summarized as follows:
\begin{itemize}[leftmargin=2em, noitemsep]
\item \textbf{An underestimated issue}: Imperfect decomposition results in ICR, causing deviations in illumination and reflectance components from their ideal states, ultimately degrading synthesized image quality.  
\item \textbf{An innovative solution}: ICR could be identified and mitigated by leveraging feature similarities between illumination and reflectance components.  
\item \textbf{A novel LLIE framework}: We propose an inter-correction Retinex framework that addresses ICR issue during both decomposition and enhancement stages.  
\end{itemize}
% \item To mitigate decomposition ICR, we design a novel dual colorspace decomposition module to accurately decompose the illumination and reflectance components for mitigate initial ICR.
% \item To reduce estimation error caused by ICR, We propose a Component-enhance and Residual-mitigate Module that produces super-resolution features for finer details and mutually mitigate the ICR in both components with information from each other.
% \begin{itemize}
% \item We pioneer an innovative approach to improve the classic Retinex model by explicitly modeling and mitigating the inter-component residual during the decomposition and enhancement process. 
% \item We design a novel enhancement unit that mutually mitigates residual of different components based on the illumination and reflectance features.
% \item We overcome the limitations of original resolution by employ  super-resolution feature enhancement to capture finer boundaries and restore lost details.
% \end{itemize}

Extensive experiments demonstrate that IRetinex outperforms state-of-the-art methods in both qualitative and quantitative evaluations. Moreover, it satisfies the requirements of computer vision applications for discriminative feature representation, achieving multimodal perception optimization.

\section{Relate works}

\textbf{Traditional Retinex-based LLIE methods.}
Traditional Retinex-based LLIE methods are non-learning approaches that rely on ideal assumptions~\cite{Guo_LIME,Jobson_multiscale_retinex,Jobson_Properties_retinex,Ng_SIAM,kimmelVariationalFrameworkRetinex2003a} or human prior knowledge~\cite{Fu_2016_CVPR,guoLIMELowLightImage2017a,Hao_Semi_Decoupled,Li_Structure} to adjust image brightness and contrast. Some methods assume smooth light distribution across images and adjust estimated illumination to enhance brightness~\cite{kimmelVariationalFrameworkRetinex2003a,Fu_2016_CVPR}, while others enhance local contrast and color fidelity by assuming that physical properties remain invariant under different lighting conditions~\cite{Jobson_multiscale_retinex,Li_Structure}. However, their dependence on hand-crafted priors limits adaptability to real-world lighting variations.

\noindent\textbf{Learning-based LLIE methods.}
Compared with traditional Retinex based methods, learning-based methods exhibit adaptability to diverse lighting scenes~\cite{loreLLNetDeepAutoencoder2017,wangLowLightImageEnhancement2022a,jiangEnlightenGANDeepLight2021}, which could be categorized into direct mapping methods and Retinex-based learning methods.

\noindent\textit{Direct mapping methods,} aim to establish a image-to-image mapping from low-light to normal-light conditions. LL-Net makes the first attempt in this trail by developing a deep auto-encoder for contrast adjustment and noise removal~\cite{loreLLNetDeepAutoencoder2017}. Subsequently, a series of networks towards this goal are proposed~\cite{renLowlightImageEnhancement2019a,luTBEFNTwobranchExposurefusion2020,xuLearningRestoreLowLight2020}. Several works~\cite{xuLowLightImageEnhancement2023a, renLowLightImageEnhancement2019b, zhuEEMEFNLowLightImage2020} employ wavelet transforms, gradient computations, and edge detectors for sharp and realistic structure enhancement. In~\cite{zhangDeepColorConsistent2022,yangAdaIntLearningAdaptive2022}, color correction is performed through the utilization of color histograms and 3D lookup tables. To capture the statistical characteristics of visual signals, Li~\etal formulate light enhancement as a task of image-specific curve estimation~\cite{liLearningEnhanceLowLight2022}. Recently, transformer-based methods with the guidance of signal-to-noise-ratio prior~\cite{xuSNRAwareLowlightImage2022a} and Fourier frequency information~\cite{wangFourLLIEBoostingLowLight2023} also exhibit accurate and visually appealing results. Despite their success, image degradation is complex and uncertain (\eg, light enhancement, color distortion and structural blurring, etc.). Directly learning the image-to-image mapping to address these varied degradations is intricate and challenging~\cite{liLowlightImageVideo2021}.
% some methods~\cite{guoLIMELowLightImage2017,fuWeightedVariationalModel2016,wangNaturalnessPreservedEnhancement2013,yingNewLowLightImage2017} employ convolutional networks for local perception, modeling color, texture, and structural relationships between pixels, \bl{and directly adjust the brightness and contrast of low-light images.} To capture long-range global contextual information, some methods~\cite{xuSNRAwareLowlightImage2022a,zhangGeneralLowLightRaw2023} integrate transformer architectures, learning global color, texture, and light information and enhancing detail recovery capabilities. With the emergence of diffusion models, some methods~\cite{wangExposureDiffusionLearningExpose,Nguyen_2024_WACV} \bl{further improve the direct learning ability by generating high-quality image details. 

\noindent\textit{Retinex-based learning methods.} Different from direct mapping methods, Retinex-based learning methods~\cite{yiDiffRetinexRethinkingLowlight,caiRetinexformerOnestageRetinexbased} have garnered significant attention due to their exceptional interpretability, as evidenced by Eq.~\eqref{eq:retinex}. Fu \etal employ convolutional networks for reflectance texture recovery while employing transformers to model global illumination dependencies~\cite{fuYouNotNeed2023}. Yi \etal  leverage diffusion models for light adjustment in illumination and detail enhancement in reflectance ~\cite{yiDiffRetinexRethinkingLowlight}. Since Zhang~\etal demonstrate the efficacy of global illumination components in guiding image reconstruction~\cite{zhangKindlingDarknessPractical2019}, a series of illumination-guided low-light enhancement schemes are proposed~\cite{wangUnderexposedPhotoEnhancement2019,caiRetinexformerOnestageRetinexbased}. 
% For instance,  Zhang \etal \cite{zhangKindlingDarknessPractical2019} introduce KinD, utilizing the illumination component to enhance the light insensitivity of the reflectance component.  
Ma~\etal perform reflectance division utilizing  illumination~\cite{maBilevelFastScene2023}. Cai~\etal design an illumination-guided multi-head self-attention that utilizes illumination deep features to guide the computation of the mapping attention~\cite{caiRetinexformerOnestageRetinexbased}. To achieve image-to-illumination mapping, Wang~\etal introduce intermediate illumination with specialized loss functions to optimize the mapping process~\cite{wangUnderexposedPhotoEnhancement2019}. These approaches effectively advance the  Retinex model performance in LLIE. However, we observe that existing methods typically assume that the illumination and reflectance is mutually independent. In real-world scenarios, the decomposition process inevitably suffers from the ICR issue, which causes the enhanced components to deviate from their ideal outcomes. This deviation ultimately degrades the quality of the final synthesized image. Therefore, it is very important to mitigate the impact of ICR to further boost the performance of Retinex models in LLIE.

% \bl{Although the above methods effectively handle different degradations by separately constructing feature mappings for illumination and reflectance, we found that the ICR associated with Retinex decomposition affects the accuracy of these mappings, resulting in a large gap between the final synthesized image and the real image. To the best of our knowledge, we are the first to explore the ICR issue. By mitigating the impact of ICR, we aim to further boosting the performance of Retinex model in LLIE.}
% treating ICR as the bias between low-light and normal-light components without considering its true fundamental properties (\ie, light residual or texture residual). 

\begin{figure*}[!t]
  \centering
    \includegraphics[width=0.98\textwidth]{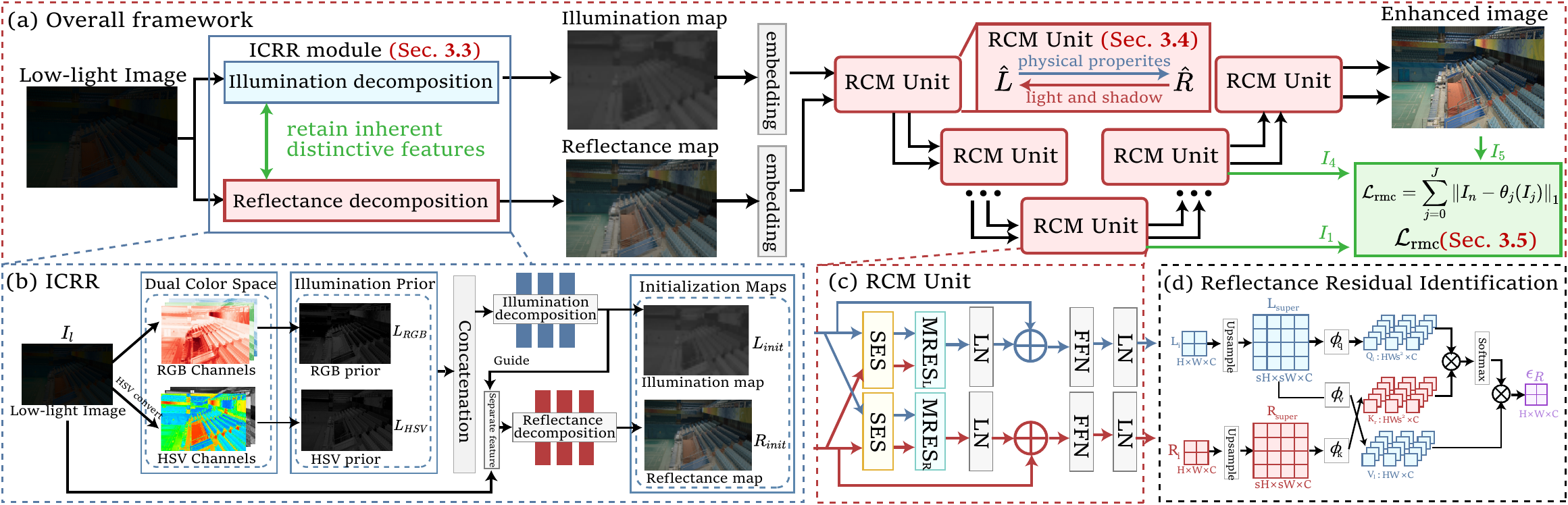}
      \vspace{-0.2cm}
  \caption{Overall architecture of IRetinex. The IRetinex consists of inter-component residual reduction module (ICRR), residual mitigation and component enhancement module (RCM) and Retinex-based multi-scale consistency loss ($\mathcal{L}_\text{rmc}$). In each RCM unite, dual-component mutually identify ICR and mitigate the accumulation of the estimation error.}
  \label{fig:framework}
  \vspace{-0.2cm}
\end{figure*}

\vspace{-0.1cm}
\section{Method}\label{Method}

In this section, we mathematically prove the impact of ICR  and introduce our  overall framework: 1) inter-component residual reduction module, 2) residual mitigation with component enhancement module, and 3) Retinex-based multi-scale consistency loss.

\subsection{Problem analysis of ICR}\label{Definition}

Based on the Retinex theory~\cite{RetinexTheory}, a low light image ${{I}_{l}}$ could be decomposed into illumination ${{L}_{l}}$ and reflectance ${{R}_{l}}$. As mentioned in Sec.~\ref{Introduction}, achieving perfect decomposition is quite challenging. {We assume that the residuals remaining in the illumination component is $\epsilon_R$ and remaining in the reflectance component is $\epsilon_L$.}
\begin{equation}
L_l = \hat{L}_l + \epsilon_R,
\label{con:deviations_l}
\end{equation}
\begin{equation}
R_l = \hat{R}_l + \epsilon_L,
\label{con:deviations_r}
\end{equation}
where $\hat{L}_l$ and $\hat{R}_l$ represent the ideal residual-free components. When enhancing these components utilizing functions $f_l(\cdot)$ and $f_r(\cdot)$, the residuals introduce estimation errors $\delta_L$ and $\delta_R$, resulting in:
\begin{equation}
L_\text{en} = f_l(L_l) = \hat{L}_\text{en} + \delta_{L},
\label{con:enhance_unit_L}
\end{equation}
\begin{equation}
R_\text{en} = f_r(R_l) = \hat{R}_\text{en} + \delta_{R}.
\label{con:enhance_unit_R}
\end{equation}

When synthesizing the enhanced image by combining these components, we get:
\begin{equation}
\begin{aligned}
I_\text{en} &= \left(\hat{L}_\text{en} + \delta_L\right) \odot \left(\hat{R}_\text{en} + \delta_R\right) \\
&= \hat{I}_\text{en} + \underbrace{\hat{L}_\text{en} \odot \delta_R + \delta_L \odot \hat{R}_\text{en} + \delta_L \odot \delta_R}_{\text{estimation error}},
\end{aligned}
\label{con:sys}
\end{equation}
where $\odot$ denotes element-wise multiplication, and $\hat{I}_\text{en}$ represents the ideal enhancement outcome.
If we express the deviation between the estimated and ideal image as $E$ (where $I_\text{en} = \hat{L}_\text{en} \odot \hat{R}_\text{en} + E$), then $E = \hat{L}_\text{en} \odot \delta_R + \delta_L \odot \hat{R}_\text{en} + \delta_L \odot \delta_R$. Through algebraic manipulation, we can derive:
\begin{equation}
\delta_L = \frac{E - \delta_R \odot \hat{L}_\text{en}}{R_\text{en}},
\label{con:residual_L}
\end{equation}
\begin{equation}
\delta_R = \frac{E - \delta_L \odot \hat{R}_\text{en}}{L_\text{en}}.
\label{con:residual_R}
\end{equation}

Assuming $E \sim \mathcal{N}(0, \sigma^2)$, $\delta_R \sim \mathcal{N}(0, \sigma^2)$, and $\delta_L \sim \mathcal{N}(0, \sigma^2)$, we observe that the distribution of $\delta_L$ is strongly related to $R_\text{en}$, and similarly, $\delta_R$ is strongly related to $L_\text{en}$.  Therefore, \BT{the estimation errors caused by ICR could be learned through a mutual correction way between illumination and reflectance components.} This crucial observation motivated the design of our IRetinex.

% Based on Eqs.~\eqref{con:accumulate_l}, ~\eqref{con:accumulate_r} and~\eqref{con:sys}, through simple algebraic operations, $\xi_L(n)$ and $\xi_R(n)$ can be expressed as:
% \begin{equation}
% \xi_L(n) = \frac{I_n - \hat{L}_n \odot R_n}{R_n},
% \label{con:residual_L}
% \end{equation}
% \begin{equation}
% \xi_R(n) = \frac{I_n - L_n \odot \hat{R}_n}{L_n}.
% \label{con:residual_R}
% \end{equation}

% {We found that illumination estimation error ($\xi_L(n)$) is strongly correlated with the reflectance component ($R_n$), and vice versa. Therefore, the estimation errors caused by ICR could be learned with the mutual assistance of different components. This crucial observation motivated the design of our IRetinex.}

\subsection{Our framework}\label{Objective}
As noted in Section~\ref{Introduction}, minimizing the initial ICR during decomposition and reducing its estimation errors in enhancement are critical. \\
\noindent \textbf{In the decomposition stage}, we aim to preserve the unique features of illumination and reflectance components. It is widely recognized that the HSV color space separates illumination from color and saturation~\cite{DouDualColorSpace2021}. Therefore, the HSV color space provides a relatively pure illumination prior, which is beneficial for mitigating residuals in the illumination component. Based on this fact, we propose an inter-component residual reduction module (Sec.~\ref{Colorspace}) to initialize the illumination and reflectance components with the assistance of HSV priors.\\
\textbf{In the enhancement stage,} driven by Eqs.~\eqref{con:residual_L} and~\eqref{con:residual_R}, and our observation that ICRs visually resemble the features of their corresponding components, we could leverage feature similarity for component's mutual correction. Specifically, we propose a mutual residual estimation module (Sec. \ref{ICR Estimation}) to identify similar features (\ie, ICRs) between components, then integrate illumination and reflectance residuals into their corresponding components, thereby driving the components toward their ideal states.
Furthermore, to recover detail losses caused by ICR, we design a {super-resolution enhancement module} (Sec.~\ref{super}) to improve the sensitivity to texture detail. Additionally, we propose a {Retinex-based multi-scale consistency loss} (Sec.~\ref{loss}) to constrain ICR estimation.

\subsection{Inter-component residual reduction module}\label{Colorspace}

% \begin{figure}[htpb]
%   \centering
%     \includegraphics[width=0.95\columnwidth]{Figures/decomposition.pdf}
%   \caption{Overall architecture of dual colorspace decomposition module. The network initializes the illumination prior from the RGB and HSV color spaces, and then utilizes deep network to estimate the illumination component, which subsequently guides the estimation of the reflectance component.}
%   \label{fig:decomposition}
% \end{figure}

As illustrated in Fig.~\ref{fig:framework} (b), our decomposition process for a low-light image $I_l$ consists of two key stages: (1) illumination initialization and (2) reflectance initialization.\\
\textbf{Illumination initialization.} We estimate the illumination component through a hybrid approach combining RGB and HSV color spaces: 1) RGB-space initialization: Following conventional practice~\cite{caiRetinexformerOnestageRetinexbased}, we compute the mean across RGB channels to establish the initial RGB illumination prior $L_{\text{rgb}} = \frac{1}{3} \sum_{c \in \{r,g,b\}} I_\text{rgb}^{(c)}$. 2) HSV-space initialization: We convert the low-light input image \( I_l \) to HSV color space and extract the V channel as an auxiliary illumination prior $L_{\text{hsv}} = I_{\text{hsv}}^{(v)}$. For illumination map decomposition, we concatenate the input features ${{I}_{l}}$, ${{L}_\text{RGB}}$, and ${{L}_\text{HSV}}$ along the channel dimension, applying convolution to adaptively separate light and shadows in low light image. Recognizing that large-scale convolutions are effective in modeling interactions between regions with varying illumination~\cite{caiRetinexformerOnestageRetinexbased}, we employ a large-scale convolution (\ie, kernel size $5\times5$) to capture light variation. The final illumination prior \({L}_\text{init} \) is formulated as:
\vspace{-0.1cm}
\begin{equation}
{{L}_\text{init}}=\psi_{1\times 1}(\psi_{5\times 5}(\psi_{1\times 1}(\text{concat}(I_l,{L}_\text{RGB},{L}_\text{HSV})))),
\label{con:objective}
\end{equation}
where $\psi_{1\times 1}$ indicate convolution operation, $\text{concat} (\cdot )$ represent concatenation along the channel dimension.

\begin{figure}[htpb]
\vspace{-0.3cm}
\centering
\includegraphics[width=0.95\columnwidth]{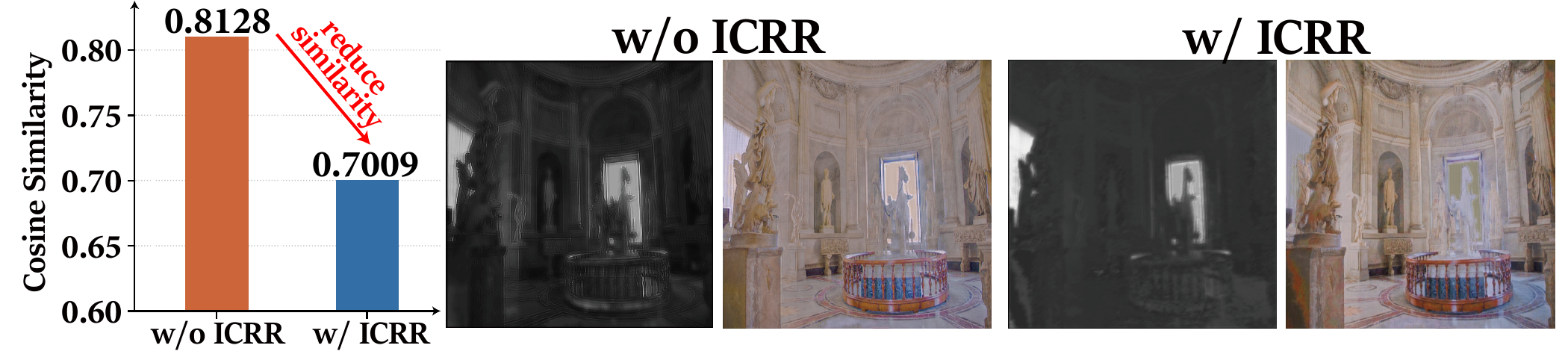} % Reduce the figure size so that it is slightly narrower than the column. Don't use precise values for figure width.This setup will avoid overfull boxes.
\vspace{-0.3cm}
\caption{ICRR reduces the feature similarity between illumination and reflectance, preserving the inherent discriminative features of each component.}
\vspace{-0.2cm}
\label{fig:simility}
\end{figure}

\noindent\textbf{Reflectance initialization.}
For reflectance component decomposition, according to Retinex theory~\cite{RetinexTheory}, the reflectance component  ${{R}_{\text{init}}}$ could be obtained  through
${{R}_{\text{init}}}={{I}_{l}}/{{L}_{\text{init}}}$, 
where $/$ denotes the element-wise division. However, this simplistic and rudimentary decomposition typically results in color distortions~\cite{Wu_URetinex_plus}. To mitigate these distortions and preserve the inherent distinctive features of reflectance components, we employ convolutional neural network to adaptively separate ${{R}_{\text{init}}}$ from ${I}_{l}$:
\begin{equation}
R_{\text{init}} \!=\! \psi_{1 \times 1} \!\left( \psi_{5 \times 5} \!\left( \psi_{1 \times 1} \!\left( \text{concat} \!\left( I_l, \text{softmax} \!\left( \frac{I_l}{L_{\text{init}} \!+\! \epsilon} \right) \right) \right) \right) \right),
\label{con:objective}
\end{equation}
where \(\text{softmax}(\cdot)\) denotes the softmax function, which normalizes the reflectance prior into a probability distribution, \(\epsilon\) is a small constant to prevent division by zero.
As illustrated in Fig.~\ref{fig:simility}, ICRR reduces the inter-component similarity between decomposed illumination and reflectance images. Specifically, it minimizes texture residuals in the illumination component and eliminating light feature in the reflectance component.
% Due to the absence of ground truth illumination and reflectance components, in IRetinex, we reconstruct enhanced images based on the Retinex theory and utilize normal light image ${I}_{n}$ for supervision. This can be simply expressed as follows:
% \begin{equation}
% {R}^\ast_\text{en},{L}^\ast_\text{en} =\underset{R_{\text{en}}, L_{\text{en}}}{\arg\min} \left\| I_{\text{n}} - R_{\text{en}} \odot L_{\text{en}} \right\|_1,
% \label{con:objective}
% \end{equation}
% where ${{\left\| \cdot  \right\|}_{1}}$ denote ${{L}_{1}}$ loss. Under the end-to-end optimization, ${R}_\text{en}$ and ${L}_\text{en}$  can gradually approach the ideal decomposition states (\ie, ${R}^\ast_\text{en}$ and ${L}^\ast_\text{en}$), thereby reducing the initial ICR in decomposition.
\vspace{-0.2cm}
\subsection{Residual mitigation and component enhancement module}\label{PDEM}
As illustrated in Fig.~\ref{fig:framework}, our enhancement network adopts a multi-scale architecture based on the residual mitigation and component enhancement module (RCM), which consists of two key schemes:  1) a mutual residual estimation scheme (MRES) to simultaneously identify and mitigate ICRs across components, and 2) a super-resolution enhancement scheme  (SES) that recovers buried details. Given the illumination and reflectance features (${{L}_{i}}$ and ${{R}_{i}}$) in layer $i$ , the enhanced features by RCM (${{L}_{i+1}}$ and ${{R}_{i+1}}$) can be expressed as:
\begin{equation}
{{L}_{i+1}}\!=\!\text{FFN}(\text{MRES}(\text{SES}({{R}_{i}},{{L}_{i}}))+{{L}_{i}}),
\label{con:r1}
\end{equation}
\begin{equation}
{{R}_{i+1}}\!=\!\text{FFN}(\text{MRES}(\text{SES}({{L}_{i}},{{R}_{i}}))+{{R}_{i}}),
\label{con:r2}
\end{equation}
where Feed-Forward Network (FFN) consists of three convolutional layers with GELU activation. Layer Normalization is omitted for clarity. Below we detail MRES and SES.

\begin{figure}[!t]
\centering
\includegraphics[width=0.95\columnwidth]{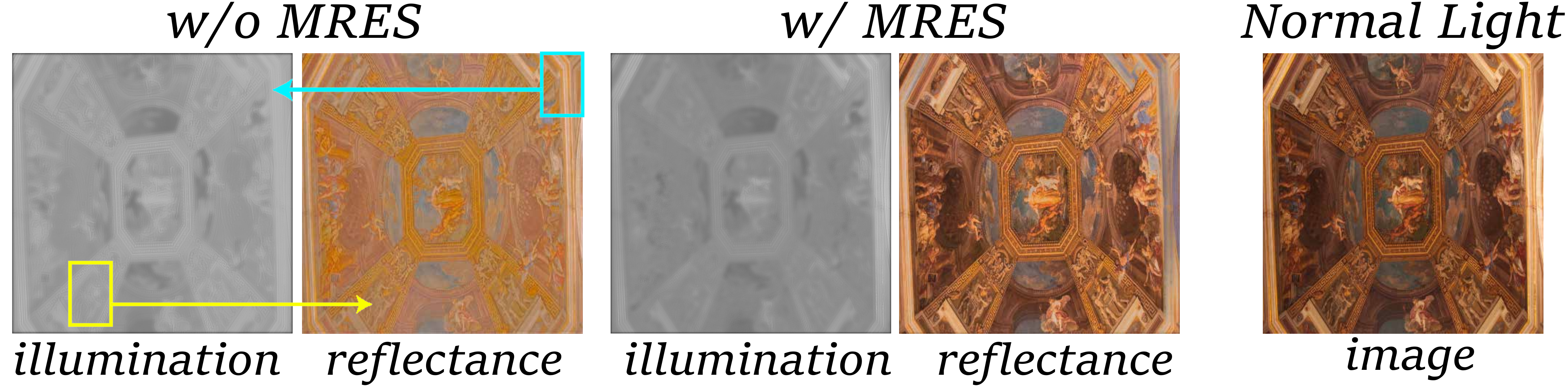}
\vspace{-0.2cm}
\caption{MRES transfers physical properties from illumination to reflectance and light features from reflectance to illumination. }
\vspace{-0.4cm}
\label{fig:reflectillu}
\end{figure}

\noindent\textbf{Mutual residual estimation scheme.} \label{ICR Estimation}
In MRES, we leverage inter-component feature similarity to capture ICR, then compensate the corresponding components with the identified residuals. Taking the illumination component as an example, first, we initialize ${Q}_{r}$, ${K}_{l}$ with depth features from both component and initialize ${V}_{r}$ from reflectance component:
\begin{equation}
{{Q}_{r}}={{\phi }_{q}}({{R_{i}}}),
{{K}_{l}}={{\phi }_{k}}({{L_{i}}}),
{{V}_{r}}={{\phi }_{v}}({{R}_{i}}),
\label{con:qkv_l}
\end{equation}
where ${{\phi }_{q}}$, ${{\phi }_{k}}$, ${{\phi }_{v}}$ denote the parameter initialize layer of ${{Q}_{r}}$, ${{K}_{l}}$ and ${{V}_{r}}$, respectively. Second, we compute the similarity matrix between reflectance and illumination features by utilizing ${{Q}_{r}}$ and ${{K}_{l}}$. Subsequently, we leverage this matrix to extract similar features from the reflectance feature, thereby capturing the light and shadow residuals within the reflectance component. The estimated reflectance residual $\epsilon_L$ can be formulated as:
\vspace{-0.1cm}
\begin{equation}
\epsilon_L =\text{softmax}\left( \frac{{{Q}_{r}}K_{l}^\top}{{{{d}_{i}}}} \right){V}_{r},
\label{con:L_mmisa}
\end{equation}
where ${{d}_{i}}\in {{\mathbb{R}}}$ is a learnable parameter that adaptively scales the matrix multiplication. Similarly, we can obtain illumination residual $\epsilon_R$ in the same way:
\vspace{-0.1cm}
\begin{equation}
\epsilon_R =\text{softmax}\left( \frac{{{Q}_{l}}K_{r}^\top}{{{{d}_{i}}}} \right){V}_{l}.
\label{con:rmmisa}
\end{equation}
\indent After reshaping, reflectance residual $\epsilon_R\in {{\mathbb{R}}^{H\times W\times C}}$ and illumination residual $\epsilon_L\in {{\mathbb{R}}^{H\times W\times C}}$ are added to original inputs ${L}_{i}$ and ${R}_{i}$ utilizing Eqs.~\eqref{con:r1} and~\eqref{con:r2}. Fig.~\ref{fig:reflectillu} visualizes the benefits of mutual residual estimation. Through MRES, we  transfer  light residuals from reflectance to  illumination, and conversely, transfer  texture residuals from illumination to reflectance, thereby driving both components closer to their ideal outcomes.

\begin{figure}[!t]
\centering
\includegraphics[width=0.95\columnwidth]{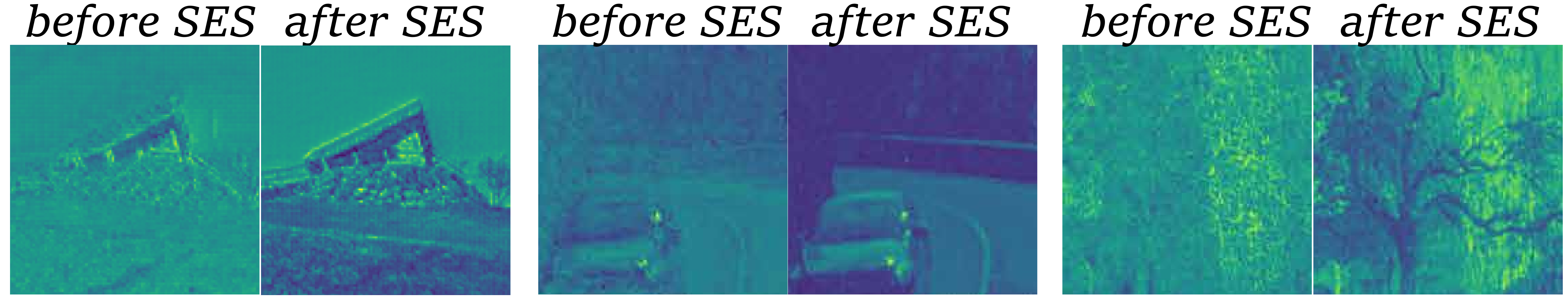} % Reduce the figure size so that it is slightly narrower than the column. Don't use precise values for figure width.This setup will avoid overfull boxes.
\vspace{-0.3cm}
\caption{SES effectively enhance the finer details of deep features.}
\vspace{-0.5cm}
\label{fig:superfeature}
\end{figure}

\noindent \textbf{Super resolution enhancement}\label{super}
SES aimed to  recovers buried details caused by ICR. Specifically, for reflectance feature ${{R}_{i}}$ and illumination feature ${{L}_{i}}$, we expand the feature resolution to initialize the high-resolution feature ${{R}_{\text{super}}}$ and ${{L}_{\text{super}}}$, represented as:

\begin{equation}
{{R}_{\text{super}}}={{\varphi }_{r}^{s}}({{R}_{i}}),{{L}_{\text{super}}}={{\varphi }_{l}^{s}}({{L}_{i}}),
\label{con:r_super}
\end{equation}
where ${{\varphi}_{r}^{s}}(\cdot)$ and ${{\varphi}_{l}^{s}}(\cdot)$ are the  upsampling networks with deconvolution layer and convolution layers, and $s$ is the scale factor. For capture illumination residual $\epsilon_R$, 
 we utilize upsampled feature ${{R}_\text{super}}$ to initializes $Q_r$ and $K_l$, the original resolution feature ${{R}_{i}}$ initializes $V_r$, Eq.~\eqref{con:qkv_l} can be rewritten as:
\begin{equation}
{{Q}_{r}}={{\phi }_{q}}({{R_\text{super}}}),
{{K}_{l}}={{\phi }_{k}}({{L_\text{super}}}),
{{V}_{r}}={{\phi }_{v}}({{R}_{i}}).
\label{con:qrkrvr}
\end{equation}
\indent Leveraging the super-resolution features, we restore details compromised by ICRs. As illustrated in Fig.~\ref{fig:superfeature}, SES refines edges, shapes, and structural details within deep features, which is beneficial in recovering textures obscured under low-light conditions. MRES and SES constitute the RCM unit. {A discussion of the complexity of RCM could be found in our supplementary material (Sec. B).}
% \begin{figure}[htpb]
%   \centering
%     \includegraphics[width=0.95\columnwidth]{Figures/RGCS.pdf}
%   \caption{Overall architecture of illumination enhance and residual mitigate process.  ${{\phi }_{q}}$, ${{\phi }_{k}}$, ${{\phi }_{v}}$ denote the parameter initialize layer of ${{Q}_{r}}$, ${{K}_{l}}$ and ${{V}_{r}}$. FC indecates fully connection layer.}
%   \label{fig:RGCS}
% \end{figure}}

\subsection{Retinex-based multi-scale consistency loss}\label{loss}
{Based on a multiscale network architecture, we could supervise ICR estimation at different scales. However, this supervision encounters two challenges:} (1) Retinex decomposition is ill-posed~\cite{zhangKindlingDarknessPractical2019}, with no explicit ground truth for reflectance and illumination components; (2) Supervising multi-scale features requires downsampling the original ground truth, potentially losing detail.
To avoid inappropriate supervision caused by ground truth decomposition, we reconstruct illumination and reflectance deep features into enhanced images:
\vspace{-0.1cm}
\begin{equation}
{{I}_{j}}={{\zeta }^{r}_{j}}({{R}_{j}})\odot {{\zeta }^{l}_{j}}({{L}_{j}}),
\label{con:multi_i}
\end{equation}
where $j\in\{0, 1, 2, 3, 4\}$ denotes the network depth, ${\zeta }^{r}{j}(\cdot)$ and ${\zeta }^{l}{j}(\cdot)$ both consist of 3×3 convolutional layers for integrating high-dimensional deep features to synthesize the enhanced image ${{I}_{j}}\in {{\mathbb{R}}^{\frac{H}{2j}\times \frac{W}{2j}\times 3}}$.

To mitigate detail loss from ground truth downsampling, we upsample different enhanced images to match the ground truth resolution. We compute pixel-wise errors utilizing $\mathcal{L}_1$ loss:
\vspace{-0.1cm}
\begin{equation}
{\mathcal{L}_\text{rmc}}=\sum\limits_{j=0}^{J}{{{\left\| {{I}_{n}}-\theta_{j}({{I}_{j}}) \right\|}_{1}}},
\label{con:rmc}
\end{equation}
where $\theta_{j}$ denotes the adaptive upsampling network for layer $j$, comprising convolutional and upsampling layers. ${\mathcal{L}_\text{rmc}}$ enables ground truth to supervise multi-scale deep feature learning and optimize illumination and reflectance features jointly.

% \subsection{Summary}
% {We now provide a summary, our IRetinex contains two stage: decomposition and enhancement.} 

% In the decomposition stage, dual color space estimation module (Sec.~\ref{Colorspace})  first decomposes the illumination component based on dual color space priors. Then, the illumination component is subsequently employed to extract the reflectance component, mitigating the initial ICR.

% In the enhancement stage, the component-enhance and residual-mitigate module (Section \ref{PDEM}) enhances illumination and reflectance components in a multual correction manner. Finally, a Retinex-based multi-scale consistency loss (Section \ref{loss}) is employed to supervise ICR estimation at each scale.

\subsection{Summary}
We now provide a summary, our IRetinex framework contains  two stages:
\textbf{In the decomposition stage,} ICRR (Section~\ref{Colorspace}) first extracts the illumination component utilizing dual color space priors. This illumination component then facilitates reflectance extraction, reducing the initial ICR.
\textbf{In the enhancement stage,} RCM (Section \ref{PDEM}) enhances both illumination and reflectance components through a mutual correction way. Finally, $\mathcal{L}_\text{rmc}$ (Section \ref{loss}) supervises ICR estimation across multiple scales.
\begin{figure*}[!t]
\centering
\includegraphics[width=0.97\textwidth]{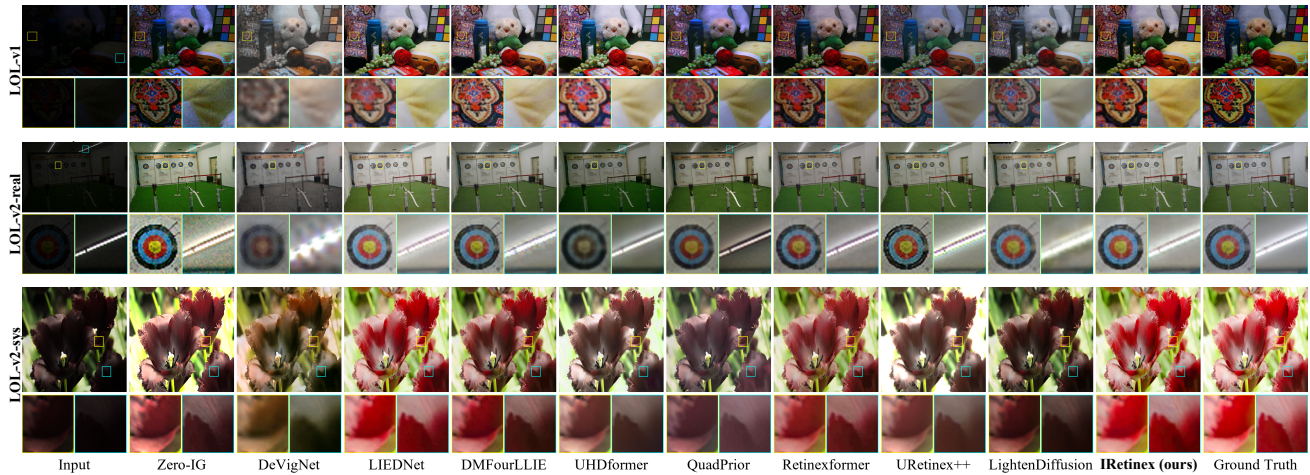} % Reduce the figure size so that it is slightly narrower than the column. Don't use precise values for figure width.This setup will avoid overfull boxes.
\vspace{-0.2cm}
\caption{Visual comparison across LOL-v1, LOL-v2-real, and LOL-v2-syn datasets. Previous methods suffer from detail loss or color distortion, whereas our algorithm effectively recovers details and maintains color fidelity.}
\vspace{-0.2cm}
\label{fig:allimage}
\end{figure*}
\section{Experiment}
% \vspace{-0.1cm}
\subsection{Datasets and implementation details}
\textbf{Datasets.} We evaluated our method on the LOL (v1~\cite{weiDeepRetinexDecomposition2018} and v2~\cite{yangSparseGradientRegularized2021}) datasets.
LOL-v1 comprises 485 pairs of low-light/normal-light images for training and 15 pairs for testing. Each pair consists of a low-light input image and a corresponding well-exposed reference image.  LOL-v2-real contains 689 pairs of low-light/normal-light images for training and 100 pairs for testing. Low-light images are mostly captured by adjusting exposure time and ISO while keeping other camera parameters fixed. LOL-v2-syn is generated by analyzing the illumination distribution in RAW format, the training and test sets are split proportionally into 900:100.

\noindent \textbf{Implementation details.}
We implement IRetinex via PyTorch. The model is trained utilizing Adam optimizer (${{\beta}_{1}}\!=\!0.9$, ${{\beta}_{2}}\!=\!0.999$) for ${1.5} \times {10}^{5}$ iterations. The learning rate is initially set to ${2} \times {10}^{-4}$ and then steadily reduces to ${1} \times {10}^{-6}$ by a cosine annealing scheme~\cite{loshchilovSGDRStochasticGradient2017} during training. The batch size is 8. The training data is enhanced by random rotation and flipping. We utilize Peak Signal-to-Noise Ratio (PSNR), Structural Similarity (SSIM), and Learned Perceptual Image Patch Similarity (LPIPS) as evaluation metrics. 
% PSNR measures the ratio between the maximum possible power of a signal and the power of corrupting noise that affects its fidelity. SSIM quantifies image similarity from the perspective of luminance, contrast, and structure, emulating human visual perception. LPIPS utilizes deep learning models to measure perceptual similarity between images.

\vspace{-0.3cm}
\subsection{Low-light image enhancement}
We evaluate our approach against 24 recent SOTA LLIE methods through qualitative and quantitative assessment.\\
\noindent \textbf{Qualitative results.} Figs. \ref{fig:allimage} present comparative results between IRetinex and existing methods on LOL-v1, LOL-v2-real, and LOL-v2-syn datasets. Enhanced images are evaluated based on color fidelity, texture preservation, and noise suppression.
The first row demonstrate texture restoration capabilities. Under low-light conditions, the textures of the pillow and toy are degraded. Although comparison methods recovered partial details, Zero-IG  exhibits significant noise compared to normal lighting images, DeVigNet shows blurred patterns on the pillow and fuzzy texture on the plush toy, while other methods suffer from varying degrees of texture loss and low color saturation. In contrast, IRetinex reconstructs texture details while maintaining color fidelity.
The second row illustrates structural recovery from underexposed images. For an archery target, competing methods exhibit noise amplification (\ie, Zero-IG, DMFourLLIE, URetinex++) and structural ambiguity (\ie, DeVigNet, LIEDNet, UHDformer). In contrast, IRetinex reconstructs the target with minimal noise and enhanced texture detail.
Moreover, when a lamp exists, simultaneously recovering both overexposed and underexposed regions is challenging. Existing methods exhibit limitations in color reproduction (\eg, UHDformer, QuardPrior) and lamp structure reconstruction (\eg, DeVigNet, DMFourLLIE, LightenDiffusion). Notably, IRetinex handles these exposure variations effectively and produces satisfactory enhancement results.
The last row presents a high-saturation image where accurate reproduction of vibrant red petal colors is challenging. Zero-IG, UHDformer, and QuardPrior exhibit color distortion. While LIEDNet and Retinexformer approximate the ground truth, they lack subtle tonal gradients within the petals. In comparison, IRetinex achieves accurate color saturation and preserving natural color transitions, maintaining fine chromatic variations.

\begin{table*}[!t]
 \caption{Quantitative comparison on LOL-v1, LOL-v2-real, and LOL-v2-syn datasets. The highest result is \textbf{bolded}, while the second highest result is \underline{underlined}.}
 \vspace{-0.2cm}
 \centering
 \small
 \setlength{\tabcolsep}{2.5mm}{
 \renewcommand{\arraystretch}{0.90}
 \begin{tabular}{c|c|ccc|ccc|ccc}
    \toprule
    \multirow{2}{*}{Methods} & \multirow{2}{*}{Venue} & \multicolumn{3}{c|}{LOL-v1} & \multicolumn{3}{c|}{LOL-v2-real} & \multicolumn{3}{c}{LOL-v2-syn} \\
    \cmidrule(lr){3-5} \cmidrule(lr){6-8} \cmidrule(lr){9-11}
    & & PSNR$\uparrow$ & SSIM$\uparrow$ & LPIPS$\downarrow$ & PSNR$\uparrow$ & SSIM$\uparrow$ & LPIPS$\downarrow$ & PSNR$\uparrow$ & SSIM$\uparrow$ & LPIPS$\downarrow$ \\
    \midrule
    RetinexNet~\cite{weiDeepRetinexDecomposition2018} & BMVC'18 & 17.54 & 0.7136 & 0.3846 & 17.65 & 0.6127 & 0.4459 & 17.30 & 0.7632 & 0.3833 \\
    KinD~\cite{zhangKindlingDarknessPractical2019} & ACM MM'19 & 19.66 & 0.7936 & 0.1577 & 15.68 & 0.7142 & 0.3964 & 16.80 & 0.6930 & 0.3609 \\
    Zero-DCE~\cite{guoZeroReferenceDeepCurve2020} & CVPR'20 & 15.04 & 0.4514 & 0.4032 & 16.94 & 0.5268 & 0.4488 & 16.25 & 0.6405 & 0.2139 \\
    MIRNet~\cite{zamirLearningEnrichedFeatures2020} & ECCV'20 & 24.13 & 0.8279 & 0.1280 & 20.44 & 0.7487 & 0.3325 & 23.41 & 0.9093 & 0.0946 \\
    DRBN~\cite{yangBandRepresentationBasedSemiSupervised2021} & CVPR'20 & 21.30 & 0.7889 & 0.2179 & 20.77 & 0.7569 & 0.2715 & 19.94 & 0.8142 & 0.2136 \\
    EnGAN~\cite{jiangEnlightenGANDeepLight2021} & TIP'21 & 17.48 & 0.6583 & 0.3142 & 18.84 & 0.6286 & 0.3061 & 16.57 & 0.7923 & 0.2212 \\
    SGM~\cite{yangSparseGradientRegularized2021a} & TIP'21 & 17.81 & 0.7648 & 0.2185 & 20.06 & 0.8191 & 0.1627 & 22.05 & 0.9031 & 0.0902 \\
    RUAS~\cite{liuRetinexinspiredUnrollingCooperative2021} & CVPR'21 & 14.63 & 0.5726 & 0.2846 & 17.97 & 0.6459 & 0.3289 & 16.94 & 0.6483 & 0.2215 \\
    Restormer~\cite{zamirRestormerEfficientTransformer2022} & CVPR'22 & 21.71 & 0.7500 & 0.2179 & 22.43 & 0.7594 & 0.2684 & 20.70 & 0.8025 & 0.1954 \\
    Uformer~\cite{wangUformerGeneralUShaped2022} & CVPR'22 & 19.20 & 0.7876 & 0.2821 & 18.60 & 0.6357 & 0.4276 & 18.21 & 0.7522 & 0.2129 \\
    SNR-Net~\cite{xuSNRAwareLowlightImage2022a} & CVPR'22 & 24.61 & \underline{0.8417} & 0.1501 & 21.48 & 0.8314 & 0.1590 & 24.13 & 0.9303 & 0.0609 \\
    SCI~\cite{maFastFlexibleRobust2022} & CVPR'22 & 17.35 & 0.6821 & 0.2077 & 16.69 & 0.5940 & 0.3480 & 17.50 & 0.7163 & 0.2171 \\
    LLFormer~\cite{liuLAENetLocallyadaptiveEmbedding2023} & AAAI'23 & 23.64 & 0.8092 & 0.1641 & 20.74 & 0.8175 & 0.1622 & 23.16 & 0.9099 & 0.0925 \\
    FourLLIE~\cite{wangFourLLIEBoostingLowLight2023} & ACM MM'23 & 20.58 & 0.6929 & 0.1646 & 22.34 & 0.8220 & \textbf{0.1135} & 24.64 & 0.9175 & 0.0690 \\
    Bread~\cite{guoLowlightImageEnhancement2023} & IJCV'23 & 18.07 & 0.7124 & 0.2436 & 18.94 & 0.7659 & 0.2598 & 17.06 & 0.8226 & 0.1918 \\
    Retinexformer~\cite{caiRetinexformerOnestageRetinexbased} & CVPR'23 & \underline{25.15} & {0.8402} & 0.1272 & 22.79 & \underline{0.8377} & 0.1691 & {25.67} & 0.9380 & 0.0639 \\
    DMFourLLIE~\cite{DMFourLLIE_zhang_2024} & ACM MM'24 & 22.43 & 0.8012 & 0.1241 & 18.86 & 0.8214 & \underline{0.1216} & 24.89 & 0.9350 & 0.0546 \\
    UHDFormer~\cite{UHDFormer_wang_2024} & AAAI'24 & 22.88 & 0.8350 & 0.1366 & 18.57 & 0.6457 & 0.4167 & 17.46 & 0.7468 & 0.2692 \\
    QuadPrior~\cite{QuadPrior_Wang_2024_CVPR} & CVPR'24 & 18.33 & 0.7675 & 0.2048 & 20.59 & 0.7600 & 0.2003 & 17.10 & 0.7457 & 0.3167 \\
    DeVigNet~\cite{luoDevignetHighResolutionVignetting2024} & AAAI'24 & 20.71 & 0.7123 & 0.2079 & 19.81 & 0.5607 & 0.4612 & 18.67 & 0.6802 & 0.3056 \\
    Zero-IG~\cite{shi_ZERO_IG_2024} & CVPR'24 & 22.71 & 0.8089 & 0.1949 & 18.13 & 0.6932 & 0.2448 & 17.43 & 0.7307 & 0.2121 \\
    LightenDiffusion~\cite{Jiang_2024_ECCV} & ECCV'24 & 19.12 & 0.7486 & 0.1954 & 21.71& 0.7889 & 0.1871 & 19.94 & 0.8263 & 0.1700 \\
    LIEDNet~\cite{LIEDNet_liu_2025_TCSVT} & TCSVT'25 & 24.79 & 0.8254 & 0.1234 &20.29&0.8181 &0.1669 & \underline{26.00} & \underline{0.9431} & \underline{0.0496} \\
    URetinex++~\cite{Wu_URetinex_plus} & TPAMI'25 & 23.83 & 0.8186 & \underline{0.1154} & \textbf{23.92} & 0.7981 & 0.1872 & 18.99 & 0.7443 & 0.2101 \\
    \midrule
    \textbf{IRetinex (ours)} & - & \textbf{25.64} & \textbf{0.8787} & \textbf{0.1107} & \underline{23.56} & \textbf{0.8472} & 0.1675 & \textbf{26.84} & \textbf{0.9513} & \textbf{0.0488} \\
    \bottomrule
  \end{tabular}
 }
 \vspace{-0.1cm}
 \label{tab:combined_results}
\end{table*}
\noindent \textbf{Quantitative results.} Tab.~\ref{tab:combined_results} presents the results on LOL-v1, LOL-v2-real, and LOL-v2-syn. The results were obtained from publications or by executing their public code. As shown in Tab.~\ref{tab:combined_results}, IRetinex demonstrates strong performance across all three datasets. Compared to the SOTA algorithms on LOL-v1 and LOL-v2-syn, IRetinex improves PSNR by 1.94\% and 3.23\%, and SSIM by 1.13\% and 1.64\%, respectively. On LOL-v2-real,  IRetinex  achieve the best SSIM (0.8472 \textit{v.s.} 0.8377) Additionally, the enhanced images from IRetinex exhibit better structure details and color fidelity than SNR-Net and URetinex++ (Fig.~\ref{fig:allimage}).
Compared to other SOTA retinex-based deep learning methods (URetinex++\cite{Wu_URetinex_plus}, LightenDiffusion~\cite{Jiang_2024_ECCV}, and Retinexformer~\cite{caiRetinexformerOnestageRetinexbased}), our IRetinex achieves SSIM improvements of 0.0601, 0.0583 and 0.0133 across the three datasets. Different from these algorithms, IRetinex reduces the inherent inter-component residuals within the Retinex model, prompting the illumination and reflection components to approach their ideal outcomes. These improvements demonstrate that mitigating inter-component residuals benefits accurate enhancement results.

\begin{table}[!t]
 \caption{Results of the ablation study. }
 \vspace{-0.2cm}
 \centering
 \tiny
 \resizebox{0.98\columnwidth}{!}{
 \renewcommand{\arraystretch}{0.9}
 \begin{tabular}{c|l|ccc}
    \toprule
    Datasets & Methods & \multicolumn{1}{l}{PSNR$\uparrow$} & \multicolumn{1}{l}{SSIM$\uparrow$} & \multicolumn{1}{l}{LPIPS$\downarrow$} \\
    \midrule
     & Ours w/o ICRR & 22.07 & 0.8509 & 0.1289 \\
          & Ours w/o MRES & 23.23 & 0.8534 & 0.1233 \\
    LOL-v1      & Ours w/o SES & 24.85 & 0.8657 & 0.1175 \\
          & Ours w/o ${\mathcal{L}_\text{rmc}}$ & 25.09 & 0.8702 & 0.1132 \\
          & \textbf{IRetinex (ours)} & \textbf{25.64}  & \textbf{0.8787} & \textbf{0.1107} \\
    \midrule
     & Ours w/o ICRR & 19.98 & 0.8265 & 0.1896 \\
          & Ours w/o MRES & 20.01 & 0.8309 & 0.1803 \\
     {LOL-v2-real}     & Ours w/o SES & 21.56 & 0.8358 & 0.1782 \\
          & Ours w/o ${\mathcal{L}_\text{rmc}}$ & 22.98 & 0.8432 & 0.1763 \\
          & \textbf{IRetinex (ours)}  & \textbf{23.56} & \textbf{0.8472} & \textbf{0.1675} \\
    \midrule
    & Ours w/o ICRR & 23.98 & 0.9102 & 0.0914 \\
          & Ours w/o MRES & 24.88 & 0.9217 & 0.0698 \\
    {LOL-v2-syn}       & Ours w/o SES & 25.64 & 0.9109 & 0.0512 \\
          & Ours w/o ${\mathcal{L}_\text{rmc}}$ & 26.01 & 0.9352 & 0.0490 \\
          & \textbf{IRetinex (ours)}  & \textbf{26.84} & \textbf{0.9513} & \textbf{0.0488} \\
    \bottomrule
    \end{tabular}%
    }%
 \label{tab:ablation}%
  % \vspace{-0.2cm}
\end{table}%

\begin{figure}[!t]
\centering
\includegraphics[width=0.95\columnwidth]{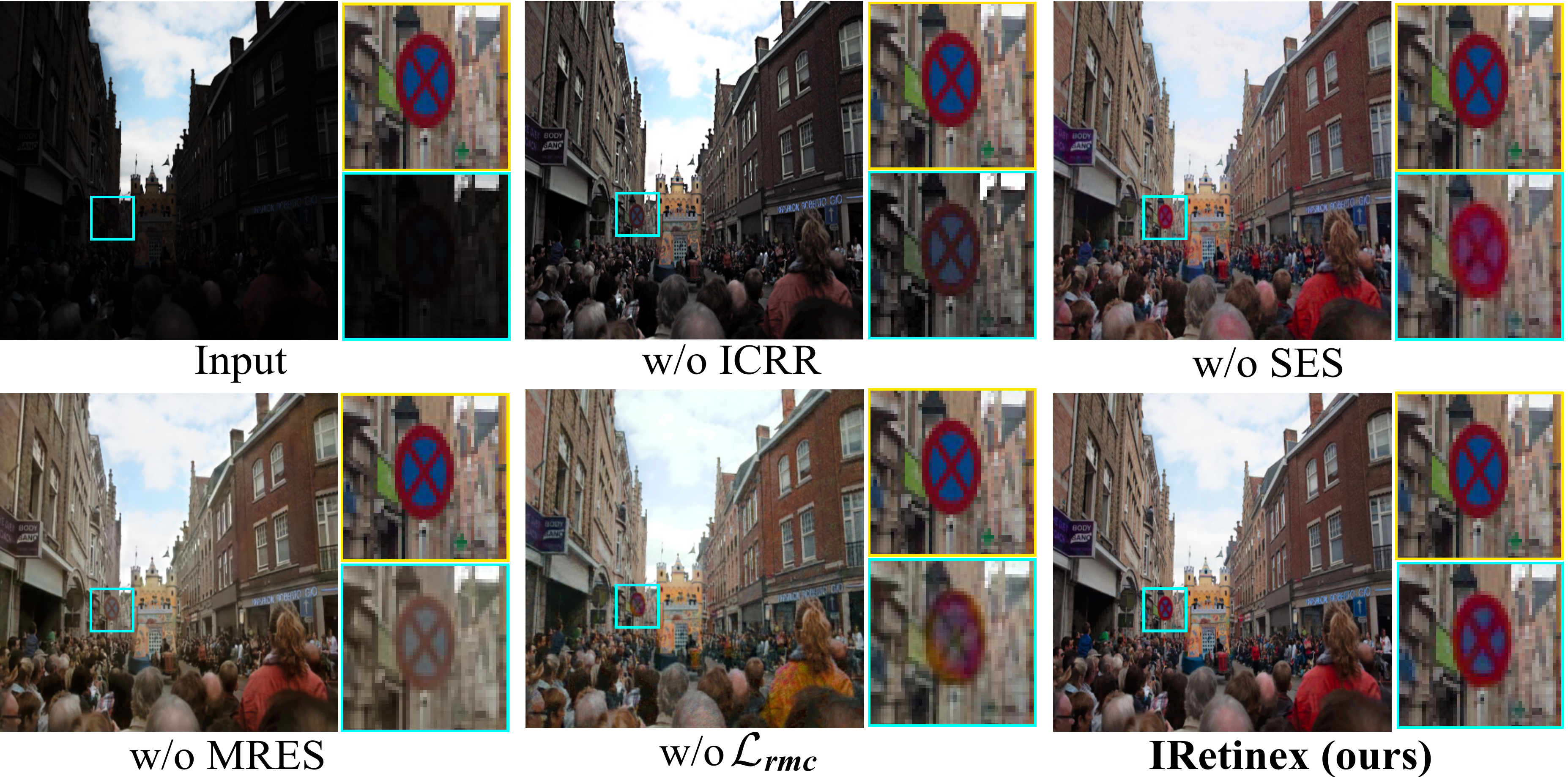} % Reduce the figure size so that it is slightly narrower than the column. Don't use precise values for figure width.This setup will avoid overfull boxes.
\vspace{-0.3cm}
\caption{Visual comparisons for ablation study on LOL-v2-syn dataset. Ground truth shown in yellow rectangles.}
\label{fig:ablation}
\vspace{-0.5cm}
\end{figure}

\begin{table}[!t]
 \caption{Ablation studies of the loss function terms. }
 \vspace{-0.2cm}
 \centering
 \tiny
 \resizebox{0.98\columnwidth}{!}{
 \renewcommand{\arraystretch}{0.9}
 \begin{tabular}{c|l|ccc}
    \toprule
    Datasets & Methods & \multicolumn{1}{l}{PSNR$\uparrow$} & \multicolumn{1}{l}{SSIM$\uparrow$} & \multicolumn{1}{l}{LPIPS$\downarrow$} \\
    \midrule
     & Ours w/o RBS & 25.12 & 0.8653 & 0.1203 \\
     {LOL-v1}     & Ours w/o ISR & 25.47 & 0.8698 & 0.1172 \\
          & \textbf{IRetinex (ours)}  & \textbf{25.64}  & \textbf{0.8787} & \textbf{0.1107} \\
    \midrule
     & Ours w/o RBS & 22.67 & 0.8404 & 0.1788 \\
     {LOL-v2-real}     & Ours w/o ISR & 23.01 & 0.8418 & 0.1704 \\
          & \textbf{IRetinex (ours)}  & \textbf{23.56} & \textbf{0.8472} & \textbf{0.1675} \\
    \midrule
     & Ours w/o RBS & 25.67 & 0.9405 & 0.0514 \\
    {LOL-v2-syn}      & Ours w/o ISR & 26.36 & 0.9499 & 0.0503 \\
          & \textbf{IRetinex (ours)}  & \textbf{26.84} & \textbf{0.9513} & \textbf{0.0488} \\
    \bottomrule
    \end{tabular}%
    }%
 \label{tab:loss}%
 \vspace{-0.8cm}
\end{table}%

\vspace{-0.2cm}
\subsection{Ablation study and further analysis}
\noindent \textbf{Component ablation.} We conducted an ablation study, with qualitative results in Fig.~\ref{fig:ablation} and quantitative results in Tab.~\ref{tab:ablation}. Without ICRR, we observed PSNR decreases of 3.57dB, 3.58dB, and 2.86dB across datasets, with enhanced images exhibiting low color saturation and indistinct structural details. Removing MRES resulted in PSNR decreases of 2.41dB, 3.55dB, and 1.96dB, though these images maintained more realistic color and clearer structure than those without ICRR. This confirms that mutual assistance between illumination and reflectance components reduces estimation error in enhancement results. Without SES, enhanced images displayed blurred structures and indistinct boundaries, with SSIM decreases of 0.0130, 0.0114, and 0.0404 across datasets. The absence of ${\mathcal{L}_\text{rmc}}$ supervision for deep features led to PSNR decreases of 0.55dB, 0.58dB, and 0.83dB, resulting in reduced brightness and blurred structures. Our complete IRetinex achieved the highest PSNR and SSIM scores, demonstrating the contribution of each component to low-light image enhancement.

\begin{figure}[htpb]
\vspace{-0.2cm}
\centering
\includegraphics[width=0.98\columnwidth]{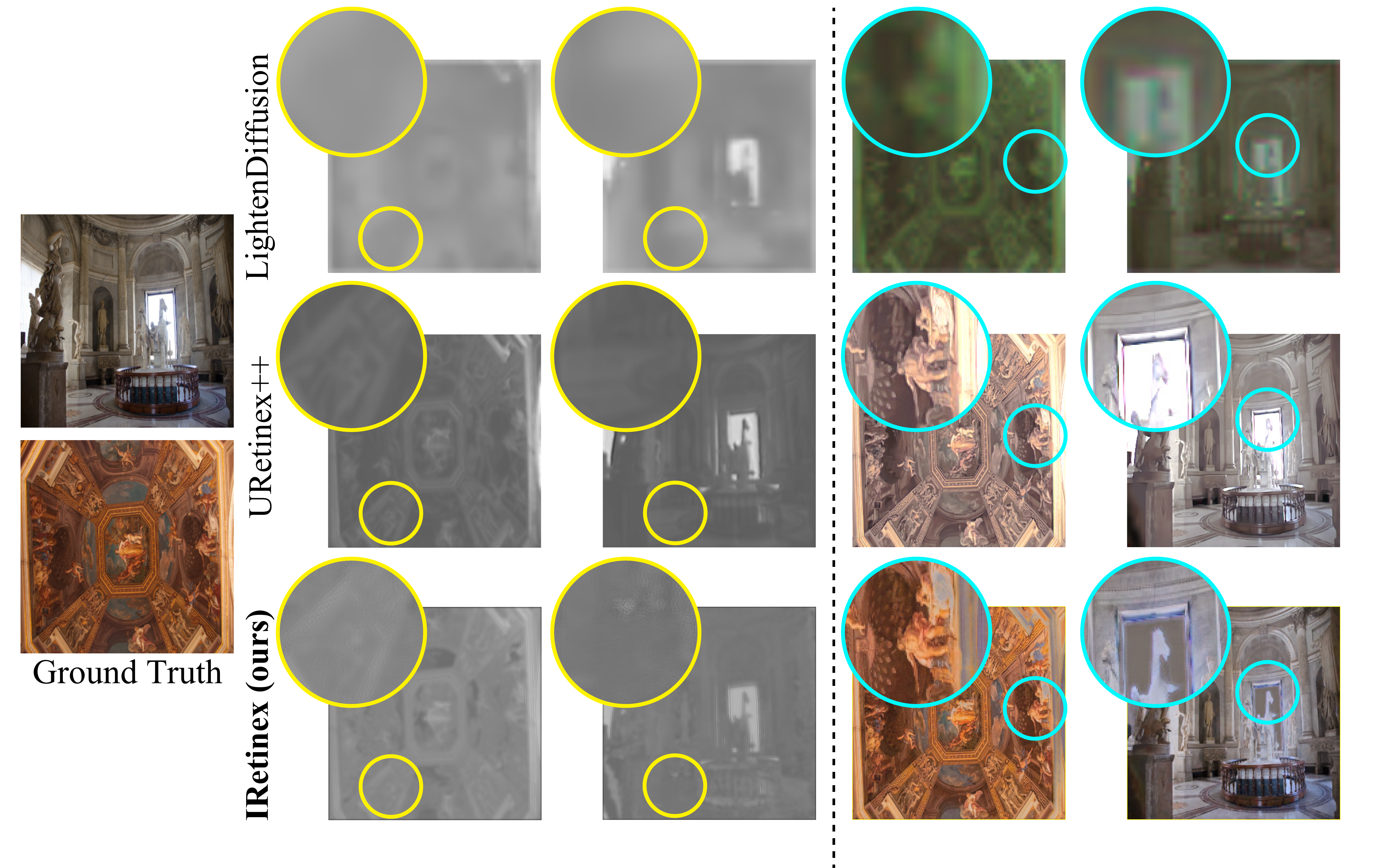} % Reduce the figure size so that it is slightly narrower than the column. Don't use precise values for figure width.This setup will avoid overfull boxes.
\vspace{-0.3cm}
\caption{Decomposition results of LLIE methods based on Retinex model on the LOL-v2-syn dataset. All illumination and reflectance components are extracted based on their original implementations.}
\label{fig:recentICRremove}
\vspace{-0.2cm}
\end{figure}

\noindent \textbf{Effectiveness of Retinex-based multi-scale consistency loss.} We conducted experiments by separately removing the image super-resolution (ISR) and Retinex-based supervision (RBS) components, with quantitative results presented in Tab.~\ref{tab:loss}. As shown in the first row, removing RBS led to performance degradation due to the inherent ill-posed nature of Retinex decomposition. The second row reveals the limitations of utilizing low-resolution images for supervision, where performance decreased across all three datasets because of detail loss in the downsampled ground truth. 

\begin{figure}[htpb]

\centering
\includegraphics[width=0.95\columnwidth]{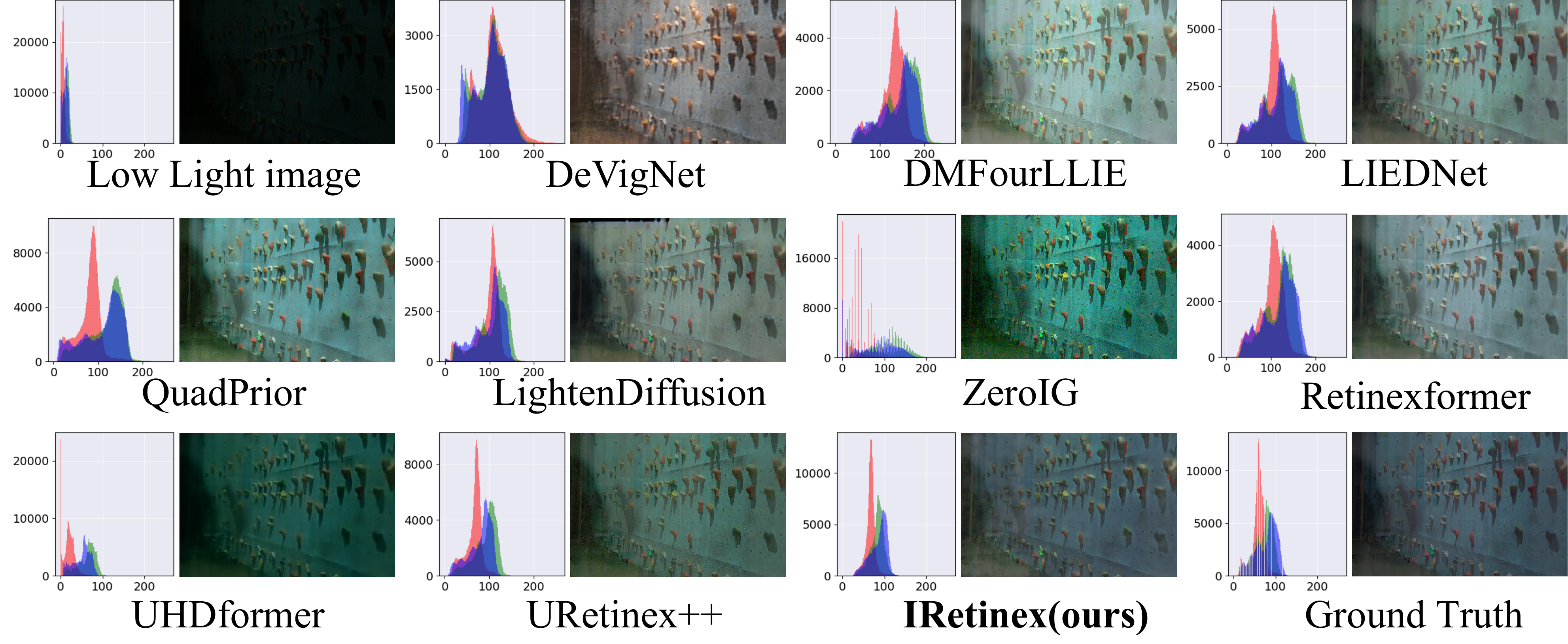} % Reduce the figure size so that it is slightly narrower than the column. Don't use precise values for figure width.This setup will avoid overfull boxes.
\vspace{-0.3cm}
\caption{RGB histogram comparison of IRetinex and other LLIE methods on the LOL-v2-real dataset. Red, green, and blue represent R, G, and B color distributions respectively. The x-axis denotes pixel values [0-255], while the y-axis indicates pixel count.}
\vspace{-0.2cm}
\label{fig:rgbhis}
\end{figure}

\begin{figure}[!t]
\vspace{-0.2cm}
\centering
\includegraphics[width=0.95\columnwidth]{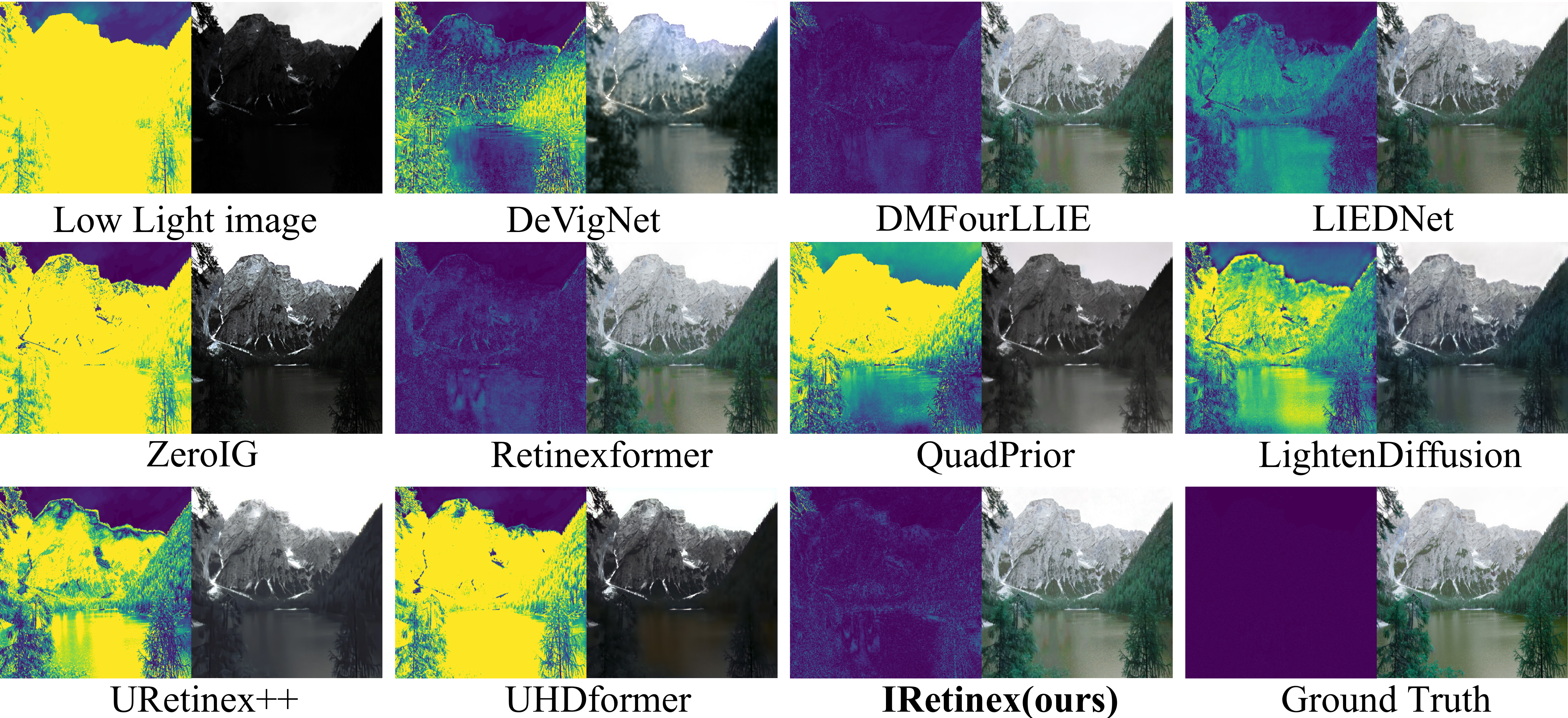} % Reduce the figure size so that it is slightly narrower than the column. Don't use precise values for figure width.This setup will avoid overfull boxes.
\vspace{-0.3cm}
\caption{Texture error map comparison between IRetinex and other LLIE methods on the LOL-v2-syn dataset. Color proximity to yellow indicates larger discrepancies, while proximity to blue denotes smaller differences.}
\vspace{-0.3cm}
\label{fig:structure_map}
\end{figure}

\begin{figure}[htpb]
\vspace{-0.2cm}
\centering
\includegraphics[width=0.96\columnwidth]{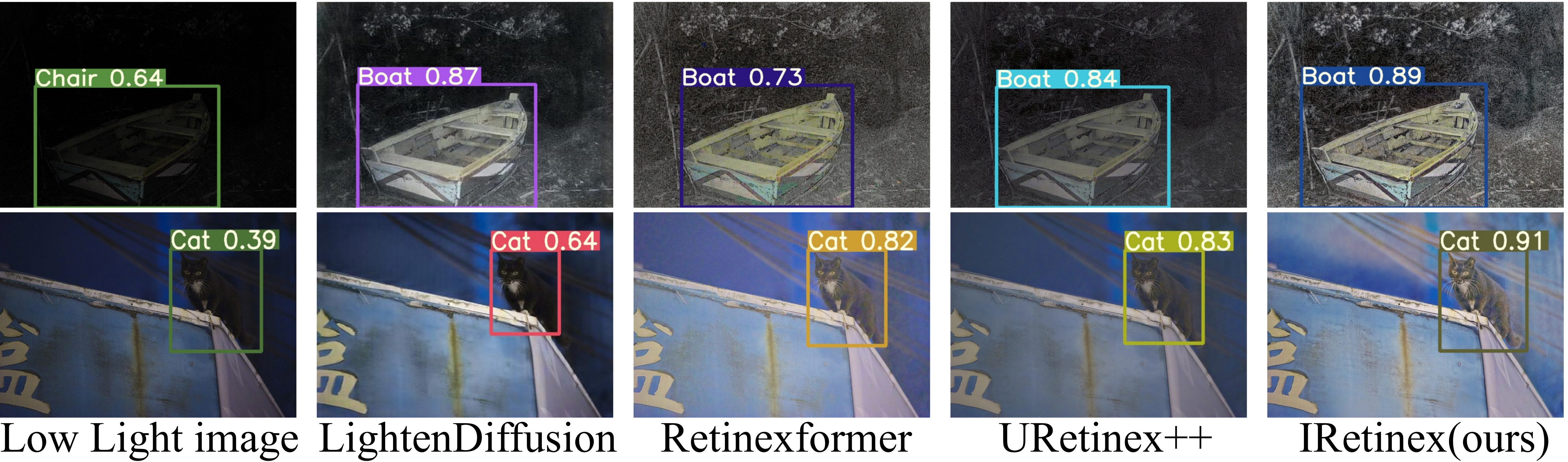} % Reduce the figure size so that it is slightly narrower than the column. Don't use precise values for figure width.This setup will avoid overfull boxes.
\vspace{-0.3cm}
\caption{Object detection performance in low-light and enhanced images on the ExDark dataset.}
\vspace{-0.3cm}
\label{fig:Exdark}
\end{figure}

\noindent \textbf{Effectiveness and benefits of ICR mitigation.} In ideal Retinex decomposition, illumination components exhibit uniform distribution across identical materials, and reflectance components accurately represent physical properties without lighting effects. Our experimental results demonstrate IRetinex's  performance in mitigating ICR compared to recent Retinex-based methods. As shown in Fig.~\ref{fig:recentICRremove}, 
URetinex++ retain textures in illumination components and display lighting residuals in reflectance components. Illumination components of LightenDiffusion are excessively smooth, with different materials exhibiting identical lighting, and the components miss textures and exhibit severe color deviation. In contrast, our method separates these components effectively. It produces illumination components with accurate light distributions and reflectance components that faithfully preserve physical details without color distortion or texture blurring. The benefits of this ICR mitigation appear in both color and structural distributions of enhanced images. RGB histograms (Fig.~\ref{fig:rgbhis}) show that IRetinex aligns with ground truth, effectively correcting the limited dynamic range characteristic of low-light images. Similarly, structural error maps (Fig.~\ref{fig:structure_map}) reveal that competing methods introduce artifacts or suffer from detail loss, whereas IRetinex minimizes structural errors and preserves fine details. These results confirm that effective ICR mitigation leads to enhanced low-light images that match normal-light references both visually and in quantitative measures. {Our supplementary materials (Sec. F, Sec. G) provide extended comparative results.}\\
\noindent \textbf{Enhancement of high-level vision tasks.}
To evaluate performance on high-level vision tasks, we applied enhanced images to challenging low-light object detection scenarios. Specifically, we trained the YOLOv7 detector~\cite{wangYOLOv7TrainableBagofFreebies2023} on the ExDark dataset~\cite{lohGettingKnowLowlight2019} using images enhanced by different LLIE methods. As illustrated in Fig.~\ref{fig:Exdark}, key object features are severely degraded under low-light conditions, bringing substantial challenges for object detection. While existing LLIE methods successfully recover obscured subjects and improve detection accuracy, our IRetinex-enhanced images exhibit sharper object boundaries and superior color fidelity, leading to further improvements in detection performance (\eg, increasing cat recognition precision from 0.39 to 0.91). These results demonstrate our method’s effectiveness in enhancing high-level vision tasks under extreme low-light conditions. In summary, the IRetinex-enhanced images maintain natural visual perception compatible with human vision while simultaneously satisfying the requirements of computer vision systems for robust feature representation. {Additional qualitative visualizations and quantitative analyses are provided in the supplementary materials (Sec. E).}

\section{Conclusion}

This work addresses a critical but overlooked challenge issue termed the inter-component residual (ICR), which arises from imperfect decomposition and degrades illumination and reflectance estimation. We discover that ICR could be mitigated by exploiting feature similarities between the two components and propose IRetinex, a novel inter-correction framework that mitigates ICR at both decomposition and enhancement stages. Extensive experiments demonstrate that IRetinex outperforms SOTA algorithms across three benchmark datasets. Crucially, our framework jointly optimizes for human vision and computer vision modalities, highlighting the importance of modeling intrinsic decomposition errors and offering a principled approach for advancing Retinex-based LLIE.

\begin{acks}
This work is supported by NSFC Project (62222604, 62206052), China Postdoctoral Science Foundation (2024M750424), Fundamental Research Funds for the Central Universities (020214380120, 020214380128), State Key Laboratory Fund (ZZKT2024A14, ZZKT2025B05), Postdoctoral Fellowship Program of CPSF (GZC20240252), Jiangsu Funding Program for Excellent Postdoctoral Talent (2024ZB242) and Jiangsu Science and Technology Major Project (BG2024031).
\end{acks}

\bibliographystyle{ACM-Reference-Format}
% \balance 
\bibliography{sample-base}
\clearpage
\appendix
This supplementary material presents extended technical implementations and experimental evaluations of the proposed IRetinex framework. Section \ref{Details clarification} provides supplementary explanations of key concepts. Section~\ref{Complexity analysis} conducts a computational complexity analysis of the MRES algorithm, establishing its superior efficiency over conventional global MSA methods. A comprehensive qualitative assessment across three benchmark datasets—LOL-v1, LOL-v2-real, and LOL-v2-syn—is provided in Section~\ref{qualitative results} to demonstrate the method's enhancement capabilities. Furthermore, Section~\ref{SID} specifically validates its robustness under extreme low-light conditions. Section~\ref{Object Detection} systematically examines IRetinex's applicability in low-light object detection scenarios, revealing its practical advantages. Sections~\ref{effec} and~\ref{Benefits} investigate two critical aspects: the efficacy of ICR elimination and its subsequent impact on visual quality improvement.  The document concludes with an insightful discussion in Section~\ref{Limitation}, discussing current limitations and outlining promising directions for future research.

\section{Supplementary explanations}\label{Details clarification}
\noindent\textbf{How ICR affects previous methods.}
ICR induces component entanglement that degrades enhancement quality. As demonstrated on challenging samples (LOL-v2, \texttt{r0809811bt.png}), our method maintains superior component independence and high PSNR, while prior approaches exhibit severe ICR effects—manifested as low cosine similarity—leading to suboptimal performance.
\begin{table}[htpb]
\vspace{-0.2cm}
\centering
\caption{Quantitative analysis of ICR effects in Retinex-Based methods.
Cosine similarity quantifies feature similarity between illumination and reflectance components.}
\vspace{-0.2cm}
\label{tab:ICR_comparison_transposed}
\setlength{\tabcolsep}{15pt}
 \resizebox{1.0\columnwidth}{!}{
\renewcommand{\arraystretch}{0.85}
\begin{tabular}{l|ccc}
\toprule
\textbf{Metric} & \textbf{KinD} & \textbf{URetinex\texttt{++}} & \textbf{Ours} \\
\midrule
Cosine Similarity & 0.9547 & 0.9581 & \textbf{0.8996} \\
PSNR (dB) & 17.06 & 16.77 & \textbf{30.35} \\
\bottomrule
\end{tabular}}
\vspace{-0.2cm}
\end{table}

\noindent\textbf{ICR effects under varying lighting conditions}
We adjusted lighting intensities utilizing gamma correction on the LOL-v2-synthetic dataset, obtaining bright ($\gamma$=0.7), moderate ($\gamma$=1.2), and dark ($\gamma$=1.5) variants of the dataset. Subsequently, we retrained IRetinex on LOL-v2-syn under varying lighting conditions. Table~\ref{tab:icr_illumination_robustness} demonstrates that ablating ICR correction from decomposition and enhancement stages substantially degrades performance, most severely in dark datasets.  This confirms that ICR issues critically undermine LLIE efficacy in extreme low-light conditions. Conversely, with ICR correction added, our method exhibited minimal performance fluctuations across all three datasets, confirming that ICR correction enables the model to maintain robustness under varying lighting conditions. 
\begin{table}[htpb]
\vspace{-0.2cm}
\Large
\centering
\caption{Quantitative evaluation of ICR correction under varying illumination levels on LOL-v2-syn.}
\vspace{-0.2cm}
\label{tab:icr_illumination_robustness}
 \resizebox{1.0\columnwidth}{!}{
 \renewcommand{\arraystretch}{1.0}
\begin{tabular}{c|c|cccc}
\hline
\textbf{Setting} & \textbf{Metrics} & \textbf{\makecell{Bright\\ ($\gamma=0.7$)}} & \textbf{\makecell{Moderate\\($\gamma=1.2$)}} & \textbf{\makecell{Dark\\($\gamma=1.5$)}} & \textbf{\makecell{Original\\($\gamma=1.0$)}} \\
\hline
w/o ICR correction &PSNR & 19.64& 17.02 & 15.32 & 17.32\\  
w/o ICR correction &SSIM & 0.8098& 0.7614 & 0.4637 & 0.7963\\
\hline
w/ ICR correction &PSNR & 26.96& 26.85 & 26.25 & 26.84\\  
w/ ICR correction &SSIM & 0.9513& 0.9512 & 0.9486 & 0.9513\\
\hline
\end{tabular}}
\vspace{-0.2cm}
\end{table}

\noindent\textbf{The function of feature similarity.}
In the enhancement stage, the feature similarity is utilized to further enhance the decomposition between illumination and reflectance features. During the decomposition stage, although ICRR reduces feature similarity between illumination and reflectance features, their cosine similarity remains high (0.7009, Figure 4), proving that reflectance and illumination are not fully decomposed. Therefore, in the enhancement stage, we further leverage feature similarity to first identify the ICR and subsequently transfer it to the corresponding target components for better decomposition. 

\noindent\textbf{The function of cosine similarity.}
We are the first to explore the ICR issue. Therefore, we specifically utilize cosine similarity to assess the mutual independence between illumination and reflectance features. We employ this metric because in high-dimensional feature spaces, cosine similarity is widely adopted for feature similarity measurement (D3still CVPR'24, RIM CVPR'24). Furthermore, we also utilized Pearson correlation to reevaluating the mutual independence and observe similar effect of cosine similarity that the similarity is low (0.8423 vs. 0.6887). 

\noindent\textbf{The reason for utilizing the dual color space prior.}
The illumination priors of RGB and HSV complement each other. Specifically: (1) RGB color space is widely utilized for illumination estimation~\cite{liLowlightImageVideo2021a}, preserves spectral details but is sensitive to color; (2) HSV color space is inherently resistant to color interference and has the ability to separate illumination (V) from color (H/S) but loses spectral details~\cite{DouDualColorSpace2021}; (3) Therefore, by combining dual color space priors, the illumination estimation process avoids color impact and preserves spectral details, producing relatively pure illumination and thereby reducing decomposition ICR. We also substituted HSV with YCbCr; however, the linear dependence of Y on RGB limits complementary information integration, resulting in a 0.9 dB PSNR reduction on the LOL-v1 dataset.

\noindent\textbf{The reason for utilizing multi-scale architecture.}
Our multi-scale architecture is a 5-layer U-shaped network with RCM as its basic unit. Multi-scale architectures are widely utilized in LLIE due to their ability to simultaneously capture global illumination and texture details (KinD~\cite{zhangKindlingDarknessPractical2019}, URetinex~\cite{Wu_URetinex_plus}). This ability facilitates ICR identification, thereby driving each component toward its ideal outcome. Ground truth downsampling causes irreversible detail loss (\eg, textures), limiting the model's ability to reconstruct detail features. To address this, we upsample each scaled feature map to high resolution for loss computation, thereby recovering multi-scale details layer by layer.

\noindent\textbf{Design motivation of ICRR and MRES.}
ICRR (0.0048M) and MRES (0.101M) are designed to suppression ICR generation and propagation, respactively.\\
- \textit{ICRR:} During decomposition, ICRR employs dual color space to initialize illumination. Specifically, HSV's V channel aids RGB color space separats illumination from color to reduce initial ICR (evidenced by low cosine similarity in Fig. 4, this metric measures component similarity, avoiding learned attention bias).\\
- \textit{MRES:} HSV lacks learning capability, thus unable to adaptively detect ICR for component purification. During enhancement, we design MRES (0.101M), which utilizes feature similarity to identify ICR and drives components toward ideal outcomes.  

\noindent\textbf{Differences from cross-attention.}
Our approach differs from cross-attention in two key aspects:\\
- \textit{Different goal:} The cross-attention in previous methods (e.g., KinD and Uretinex) focused on illumination-assisted reflectance enhancement without explicit component purification. We are the first to explore the inter-component residuals (ICR) problem through mutual correction manner, particularly emphasizing illumination-reflectance mutual purification.\\
- \textit{Different technique:} Different from Uretinex and Retinexformer that directly concatenate/add illumination features in their cross-attention, we employ similarity matrices to identify ICR and supplement them into corresponding component to approach ideal outcomes.  
Thus, our method and cross-attention are two distinct LLIE solutions.

\noindent\textbf{Rationale for assumptions.}
Our assumptions are consistent with prior work (KinD~\cite{zhangKindlingDarknessPractical2019}). Moreover, Gaussian and Poisson noise are widely utilized to simulate images captured in real-world low-light scenes~\cite{liLowlightImageVideo2021a}.

\section{Complexity analysis of ICR estimation} \label{Complexity analysis}
The computational complexity of our ICR estimate primarily stems from the matrix multiplications in Eqs.~\eqref{con:L_mmisa} and~\eqref{con:rmmisa}, \ie, $\mathbb{R}^{{C}\times{HWs^{2}}}\times\mathbb{R}^{{HWs^{2}}\times{C}}$ and $\mathbb{R}^{{HW}\times{C}}\times\mathbb{R}^{{C}\times{C}}$. Thus, the complexity $\mathcal{O}(\text{MRES})$ can be expressed as:
\begin{equation}
\begin{aligned}
\mathcal{O}(\text{MRES}) & =\mathcal{O}(\text{MRES}_{r})+\mathcal{O}(\text{MRES}_{l})\\
& =2\left[C\cdot\left(C\cdot HWs^{2}\right)+H W \cdot\left(C \cdot C\right)\right]\\
& =2(HWs^{2}C^{2}+HWC^{2})\\
& =2(s^2+1) H W C^{2},
\end{aligned}
\label{con:complexity}
\end{equation}
while the complexity of the global MSA (G-MSA) utilized by previous CNN-Transformer methods SNR-Net~\cite{xuSNRAwareLowlightImage2022a} is:
\begin{equation}
\mathcal{O}(\text {G-MSA})=2(H W)^{2} C.
\label{con:complexitysnr}
\end{equation}

Comparing Eq.~\eqref{con:complexity} and Eq.~\eqref{con:complexitysnr}, we note that $\mathcal{O}(\text{G-MSA})$ is quadratic with input space size ($HW$), imposing a substantial computational burden. Conversely, $\mathcal{O}(\text{MRES})$ is linear with space size. While we introduce upsampling factor $s$, the additional cost is lower than $HW$. 
For example, in the case where the input image dimension is $256\times256\times3$ , the G-MSA parameter is 2.099M, while MRES requires only 0.101M. Moreover, during inference, as demonstrated in Table~\ref{tab:inference}, IRetinex achieves SOTA performance  while utilizing only 5.37M parameters. Processing a $385\times385$ low-light image with our proposed IRetinex requires merely 0.075 seconds. This represents a 40.87\% reduction in inference time compared to SNR-Net (0.127 seconds), balancing  efficiency and  performance.

\begin{table}[htpb]
\vspace{-0.2cm}
 \caption{Inference efficiency comparison on an A800 GPU.} % Corrected typo & improved phrasing
 \vspace{-0.2cm}
 \Large
 \centering
 \setlength{\tabcolsep}{7pt}
 \resizebox{1.0\columnwidth}{!}{
 \renewcommand{\arraystretch}{1.0}
 \begin{tabular}{c|ccccc}
\hline
\textbf{Metric} & \textbf{\makecell{MIRNet\\ \cite{zamirLearningEnrichedFeatures2020}}} & 
\textbf{\makecell{SNR-Net\\ \cite{xuSNRAwareLowlightImage2022a}}} & 
\textbf{\makecell{Lighten\\Diffusion\\~\cite{Jiang_2024_ECCV}}} & 
\textbf{\makecell{LIEDNet\\ \cite{LIEDNet_liu_2025_TCSVT}}} &  % Fixed
\textbf{Ours} \\
\hline
PSNR & 23.41 & 24.13 & 19.94 & 26.00 & \textbf{26.84} \\
Model Size (M) & 31.79 & 39.12 & 27.83 & \textbf{4.76} & 5.37 \\
Inference time (s) & 0.263 & 0.127 & 0.537 & 0.084 & \textbf{0.075} \\ % Added space for consistency
\hline
\end{tabular}
    }%
 \label{tab:inference}%
 \vspace{-0.4cm}
\end{table}%

\begin{figure*}[!t]
\centering
\includegraphics[width=1.0\textwidth]{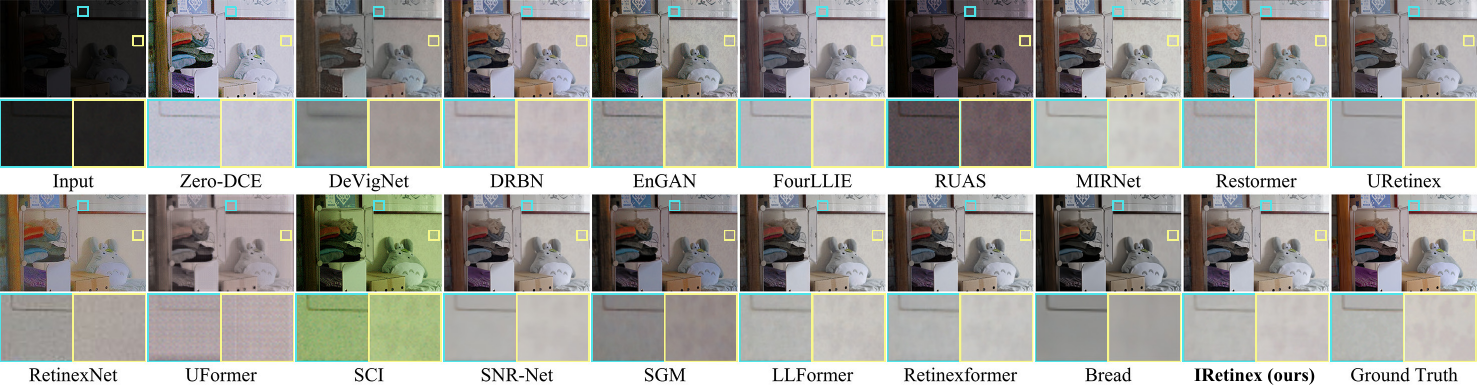} % Reduce the figure size so that it is slightly narrower than the column. Don't use precise values for figure width.This setup will avoid overfull boxes.
 \vspace{-0.4cm}
\caption{Visual comparison on LOL-v1 dataset. Previous methods collapse by nartifacts, halos, or color distortion. While our algorithm can effectively remove the noise and reconstruct high quality image details.}
\label{fig:LOLV1}
 \vspace{-0.1cm}
\end{figure*}

\begin{figure*}[!t]
\centering
\includegraphics[width=1.0\textwidth]{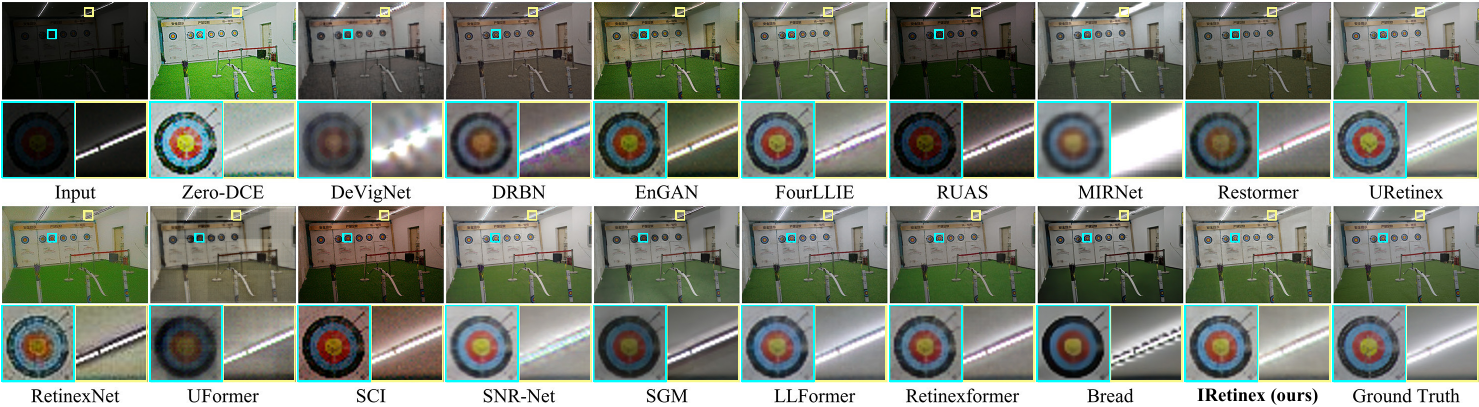} % Reduce the figure size so that it is slightly narrower than the column. Don't use precise values for figure width.This setup will avoid overfull boxes.
 \vspace{-0.4cm}
\caption{Visual comparison on LOL-v2-real dataset. Previous methods  suffer from color and structure distortion. In contrast, our algorithm minimizes noise in underexposed areas, producing reliable enhancement structure.}
 \vspace{-0.1cm}
\label{fig:LOLV2real}
\end{figure*}

\begin{figure*}[!t]
\centering
\includegraphics[width=1.0\textwidth]{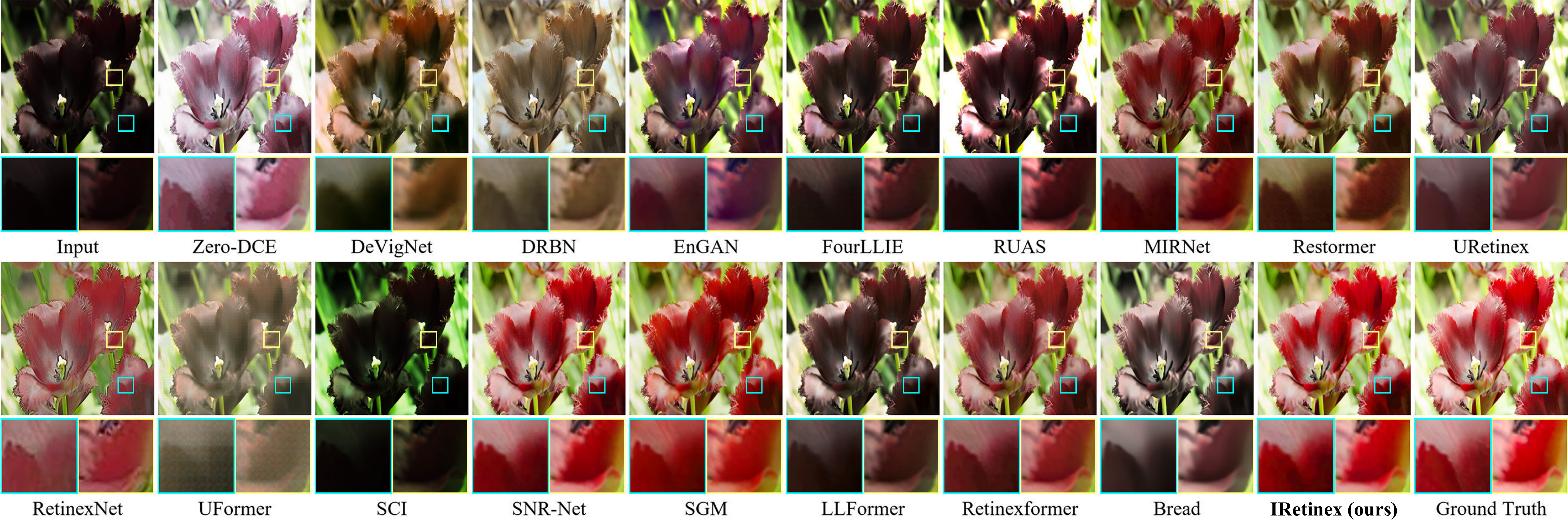} % Reduce the figure size so that it is slightly narrower than the column. Don't use precise values for figure width.This setup will avoid overfull boxes.
 \vspace{-0.4cm}
\caption{Visual comparison on LOL-v2-syn dataset. Previous methods struggle to recover the vibrant red of the petals. In contrast, our algorithm exhibits more distinct color gradations, better emulating the natural color distribution under normal light conditions.}
 \vspace{-0.1cm}
\label{fig:LOLV2sys}
\end{figure*}
\section{More qualitative results} \label{qualitative results}
Figs. \ref{fig:LOLV1}-\ref{fig:LOLV2sys} present qualitative comparisons between our IRetinex and SOTA methods on LOL-v1, LOL-v2-real, and LOL-v2-syn datasets. The enhanced images are evaluated based on three criteria: background noise, structural clarity, and color fidelity.

Fig.~\ref{fig:LOLV1} presents a comparative analysis of texture restoration results across various methods. Under low-light conditions, wallpaper and drawing paper textures are significantly degraded. Previous approaches exhibit diverse limitations: RUAS and SGM produce images with insufficient saturation, while EnGAN, RetinexNet, and Uformer amplify background noise. Although DeVigNet, URetinex, and Bread effectively mitigate noise, they introduce over-smoothing artifacts, resulting in loss of fine textures. DeVigNet additionally suffers from structural blurring.
In contrast, our proposed method effectively recovers authentic details without introducing additional noise, artifacts, and color distortions,  preserving both structural integrity and color authenticity.

Fig.~\ref{fig:LOLV2real} demonstrates the efficacy of various methods in recovering object structures from low-light images. For the underexposed archery target, competing methods exhibit different types of limitations \eg,excessive noise (RUAS, RetinexNet, SCI), structural ambiguity (DeVigNet, MIRNet, Uformer), and color distortion (Zero-DCE, DRBN, EnGAN). In contrast, our IRetinex restores the archery target with  minimal noise,  accurate color representation and finer texture details.
The right image presents a more complex scenario with simultaneous overexposed and underexposed regions. Competing methods struggle with color distortion (RetinexNet, SNR-Net) and accurate lamp structure reconstruction (DeVigNet, MIRNet, Bread). Our IRetinex method effectively handles varying exposure levels, producing reliable enhanced image.

Fig.~\ref{fig:LOLV2sys} presents an image with high color saturation, where recovering the vibrant red hues of the petals poses a significant challenge. Some methods struggle to accurately reproduce these saturated colors. For instance, Zero-DCE, DRBN, and LLFormer exhibit various degrees of color distortion, failing to capture the true vibrancy of the petals. In contrast, the enhanced images produced by SNR-Net, SGM, and our IRetinex closely approximate the ground truth. However, while SNR-Net and SGM achieve improved color saturation, they lack the nuanced color transitions within the petals that are characteristic of natural lighting conditions. IRetinex not only achieves accurate  color saturation but also exhibits distinct and natural color gradations within the petals. These results demonstrate that our method faithfully emulates the color distribution under normal lighting conditions and better recovers fine-grained color variations, contributing to the visual realism of the enhanced image.

\begin{table}[htbp]
\vspace{-0.2cm}
\caption{Quantitative comparison on SID dataset. }
\renewcommand{\arraystretch}{1.0}
\vspace{-0.3cm}
  \centering
  \Large
  \resizebox{1.0\columnwidth}{!}{
    \begin{tabular}{c|ccccc}
    \toprule[1.0pt]
       \textbf{Metrics} & \textbf{Zero-IG} & \textbf{LightenDiffusion} & \textbf{LIEDNet} & \textbf{URetinex++} & \textbf{IRetinex (ours)} \\
    \midrule
    PSNR & 17.76 & 16.65 & 21.66  & 19.12   &\textbf{25.03 }  \\  
    SSIM & 0.6483 & 0.4036 &0.6714  & 0.6022   & \textbf{0.6958}  \\
    \bottomrule[1.0pt]
    \end{tabular}}%
\vspace{-0.5cm}
\label{tab: SID}
\end{table}%

\section{SID dataset experiments}\label{SID}
Beyond standard benchmarks, we evaluated IRetinex on the subset of SID dataset~\cite{Chen_2019_ICCV}, which represent extreme low-light scenarios with near-zero illumination levels—the practical limits of visibility. It contains 2697 RAW short-/long-exposure image pairs captured by Sony $\alpha$7S II. We obtained the low-/normal-light RGB images by utilizing the same in-camera signal processing of SID to transfer RAW to RGB. For model evaluation, 2,099 pairs are designated for training and 598 for testing.
As quantified in Table~\ref{tab: SID}, IRetinex achieves state-of-the-art performance with PSNR of 25.03 dB and SSIM of 0.6958. It outperforms the second-best LIEDNet by +3.37 dB in PSNR and +0.0244 in SSIM, demonstrating exceptional robustness across diverse lighting environments.

\begin{figure*}[!t]
\centering
\includegraphics[width=0.98\textwidth]{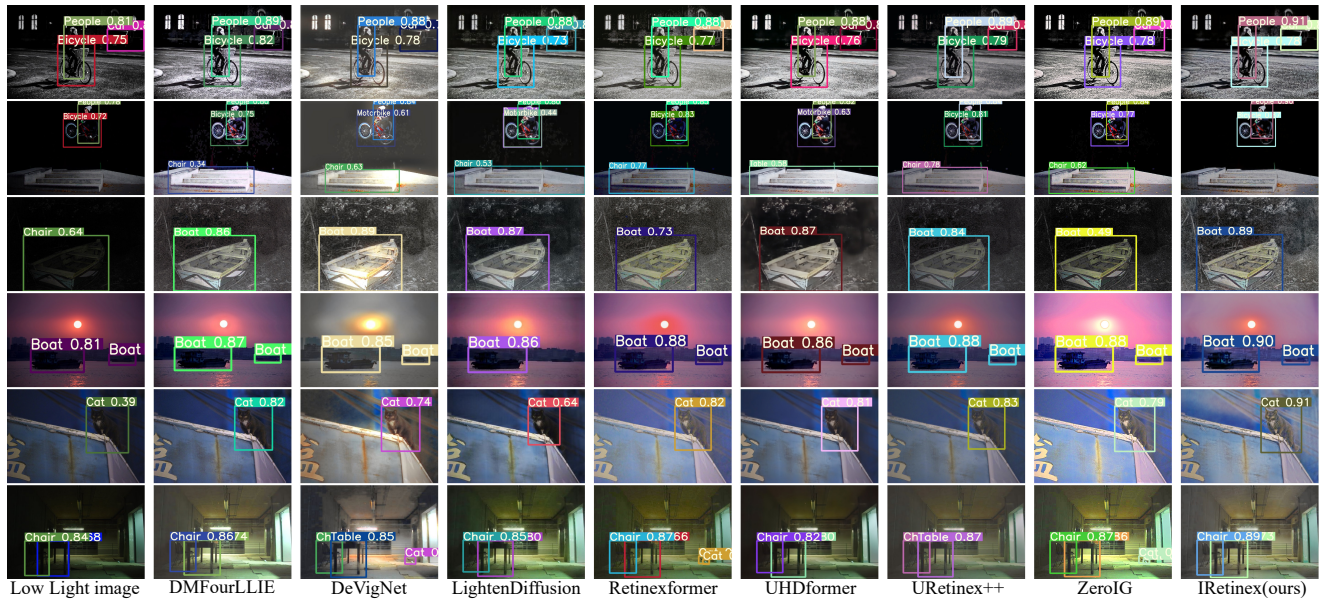} % Reduce the figure size so that it is slightly narrower than the column. Don't use precise values for figure width.This setup will avoid overfull boxes.
\vspace{-0.2cm}
\caption{Object detection performance in low-light and enhanced images on the ExDark dataset.}
\label{fig:More_Exdark}
\end{figure*}
\begin{figure*}[!t]
\centering
\includegraphics[width=0.98\textwidth]{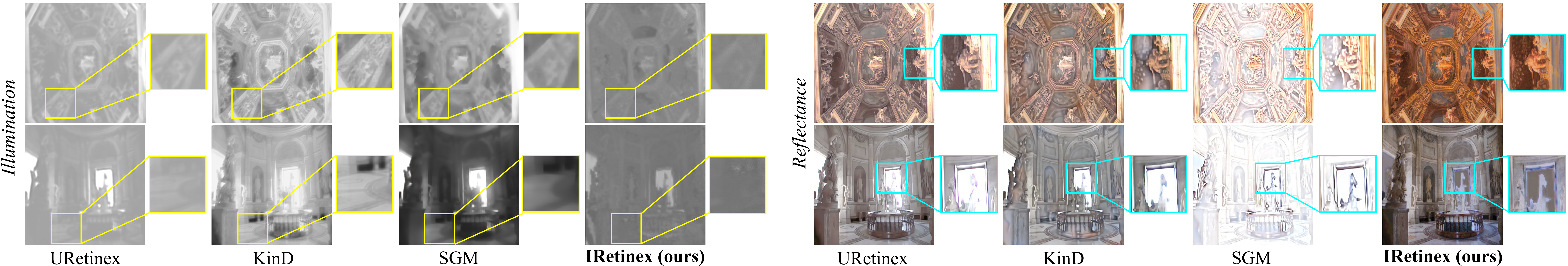} % Reduce the figure size so that it is slightly narrower than the column. Don't use precise values for figure width.This setup will avoid overfull boxes.
\vspace{-0.2cm}
\caption{Decomposition results of LLIE methods based on Retinex model on the LOL-v2-syn dataset. All illumination and reflectance components are extracted based on their original implementations.}
\label{fig:ICRremove}
\end{figure*}
\begin{table*}[!t]
\vspace{-0.2cm}
 \caption{Low-light detection results on ExDark  enhanced by different algorithms.The highest result is \textbf{bolded}, while the second highest result is \underline{underlined}.}
 \vspace{-0.1cm}
 \centering
 \tiny
 \resizebox{1.0\textwidth}{!}{
 \renewcommand{\arraystretch}{1.0}
 \begin{tabular}{l|cccccccccccc|c}
    \toprule[0.3pt]
    \rowcolor{gray!20}
    Object Categories & Bicycle & Boat  & Bottle & Bus   & Car   & Cat   & Chair & Cup   & Dog   & Motor & People & Table & Mean \\
    \midrule[0.1pt]
    DeVigNet~\cite{luoDevignetHighResolutionVignetting2024} & 83.6 & 79.2 & 81.1 & 93.7 & 86.8 & 72.6 & 69.1 & 79.7 & 80.2 & 79.8 & 82.2 & 64.0 & 79.3 \\
    DMFourLLIE~\cite{DMFourLLIE_zhang_2024} & 85.3 & 81.1 & 82.8 & \underline{94.9} & 87.7 & \underline{76.6} & 71.4 & \underline{81.4} & \textbf{83.1} & 80.9 & \underline{83.8} & 65.3 & 81.2 \\
    UHDFormer~\cite{UHDFormer_wang_2024}& 83.5 & 78.9 & 81.8 & 93.4 & 86.1 & 74.5 & 70.8 & 79.7 & 80.2 & \textbf{81.6} & 82.0 & \textbf{65.9} & 79.9 \\
    QuadPrior~\cite{QuadPrior_Wang_2024_CVPR} & 84.9 & 80.2 & 82.1 & 94.6 & \underline{87.8} & 73.9 & 71.2 & 80.9 & {82.8} & 80.1 & 83.5 & 64.7 & 80.6 \\
    Zero-IG~\cite{shi_ZERO_IG_2024}& 83.9 & 79.5 & \underline{82.9} & 94.8 & 87.2 & 75.8 & 70.8 & 80.4 & 82.3 & 79.8 & 82.1 & 63.9 & 80.3 \\
    LightenDiffusion~\cite{Jiang_2024_ECCV} & 83.7 & 80.5 & 82.7 & 93.8 & 86.3 & 74.2 & 71.8 & 80.4 & 81.5 & \underline{81.3} & 83.4 & 64.6 & 80.4 \\
    LIEDNet~\cite{LIEDNet_liu_2025_TCSVT} & 85.4 & 80.8 & 82.3 & 93.1 & 86.5 & 74.9 & \underline{72.2} & 80.8 & 81.1 & 80.9 & 83.2 & 64.3 & 80.5 \\
    URetinex++~\cite{Wu_URetinex_plus} & 85.3 & 80.9 & 82.4 & 93.6 & 87.5 & 75.5 & 71.5 & 81.1 & 82.5 & 80.8 & 83.6 & \textbf{65.5} & 80.9 \\
    Retinexformer~\cite{caiRetinexformerOnestageRetinexbased} & \underline{86.0}  & \underline{81.3}  & 81.7  & 94.3  & 87.5  & \textbf{77.3}  & \textbf{72.4}  & \textbf{82.4} & 80.8  & 80.0  & \textbf{84.5} & 64.7  & 81.1 \\
    \midrule[0.1pt]
    \textbf{IRetinex (ours)}  & \textbf{86.2} & \textbf{81.4} & \textbf{83.5} & \textbf{95.1} & \textbf{88.0}  & 76.7  & 71.7  & 81.3  & \underline{82.9}  & 80.1  & 83.8  & 64.8  & \textbf{81.3} \\
    \bottomrule[0.3pt]
  \end{tabular}
    }%
 \label{tab:ExDark}%
\vspace{-0.1cm}
\end{table*}%

\section{Low-Light object detection}\label{Object Detection}

To comprehensively evaluate low-light image enhancement performance, we applied our IRetinex to challenging high-level vision tasks, \ie, low light object detection. Speciafically, we employed the ExDark dataset~\cite{lohGettingKnowLowlight2019} to perform multi-object detection on low light images.

\subsection{Experimental setup} The ExDark dataset consists of 12 categories and contains 7,363 underexposed images, with 5,896 images utilized for training and 1,467 for testing. Various low-light enhancement methods are employed as pre-processing modules, and the YOLO-v7~\cite{wangYOLOv7TrainableBagofFreebies2023} detector is trained from scratch for object detection, utilizing average precision~(AP) as the evaluation metric.

\subsection{Qualitative results} Fig.~\ref{fig:Exdark} presents a visual comparison of object detection results in low-light scenes and their enhanced results utilizing competing methods. 
The fifth row highlights issues with bounding box accuracy, where methods like LightenDiffusion, and URetinex++ fail to fully encompass the cat's tail.  The second and sixth row showcases misclassification, with a crumpled paper erroneously identified as a cat by DeVigNet, Zero-IG, and Retinexformer.In the remaining rows, the comparative methods exhibit varying degrees of accuracy degradation relative to IRetinex, highlighting the persistent challenges in low-light image enhancement for downstream visual tasks.
In contrast, images enhanced by IRetinex enable the detector to rectify detection errors and accurately predict bounding boxes across all scenarios. This suggests that our method improves high-level visual understanding, enhancing both classification accuracy and semantic information extraction in challenging low-light conditions.

\subsection{Quantitative results} Tab.~\ref{tab:ExDark} presents the average precision (AP) for 12 categories as well as the mean AP across all categories. Our IRetinex achieved the highest mean score of 81.3 AP, surpassing the second best method DMFourLLIE by 0.1 AP and the recent fully supervised method Retinexformer by 0.2 AP~\cite{caiRetinexformerOnestageRetinexbased}. Additionally, IRetinex achieved the best results in five object categories: bicycle (86.2 AP), boat (81.4 AP), bottle (83.5 AP), bus (95.1 AP), and car (88.0 AP). Moreover, it closely approaches SOTA values in the remaining seven categories. These quantitative results demonstrate that IRetinex effectively enhances the visual quality of low-light images, ensuring the authenticity of reconstructed details and the accuracy of object information

\section{Effectiveness of ICR mitigation} \label{effec}
In an ideal ICR-free state, areas of identical materials within the illumination component should display uniform illumination distribution. Simultaneously, the reflectance component should be free from light and shadow effects, accurately representing physical properties  of the scene. To validate the ability of our method in removing ICR from different components, we compared the illumination and reflectance components with those obtained from other Retinex-based LLIE models, shown in Fig.~\ref{fig:ICRremove}. 

As illustrated in the left portion, the illumination components generated by RetinexNet, URetinex, Kind, and SGM still contain textures in regions of the same material. This indicates that physical properties (partial reflectance) are retained in the illumination component, which decreases the quality of the illumination component. In contrast, our method demonstrates fewer textural residuals and faithfully exhibits the lighting distribution of the scene.

The right figure further demonstrates the effective reduction of ICR in the reflectance components. Specifically, in the comparison methods, the reflectance components still exhibit light and shadows. The physical details of some reflectance images are disrupted by these residuals. The reflectance components also suffer from color distortion and blurred structures. In comparison, IRetinex effectively separates the illumination components, with the reflectance components faithfully representing the physical information of the scene.

\section{Benefits of ICR mitigation} \label{Benefits}
A well-enhanced low-light image should exhibit similar color and structure distributions to those of a normal-light image. Therefore, to investigate the benefits of ICR removal for image visual quality, we further compare the color and structure distributions with above  methods. The RGB histograms and structural error maps are illustrated in Fig.~\ref{fig:rgb_more} and Fig.~\ref{fig:error_more}, respectively. 

As shown in Fig.~\ref{fig:rgb_more}, low-light images exhibit a marked bias in RGB histograms, with pixel values clustered within a low dynamic range. While some methods, \eg, URetinex, partially address this bias, they still display significant pixel value discrepancies. In contrast, IRetinex closely aligns with the ground truth in both visual appearance and RGB histograms, effectively correcting color bias and producing accurate low-light images. Fig.~\ref{fig:error_more} reveals the loss of structure and texture under low-light conditions. Although the compared methods restore some details, they still suffer from structural blurring. For instance, the image produced by SGM still contains artifacts on the lake surface. In contrast, IRetinex effectively corrects the artifacts and shows minimal structural error. These results demonstrate that mitigating the inherent inter-component residuals correct both color and structural deviations, promoting the enhanced images align with normal-light images both visually and in fine details.
\begin{figure}[htpb]
\centering
\includegraphics[width=1.0\columnwidth]{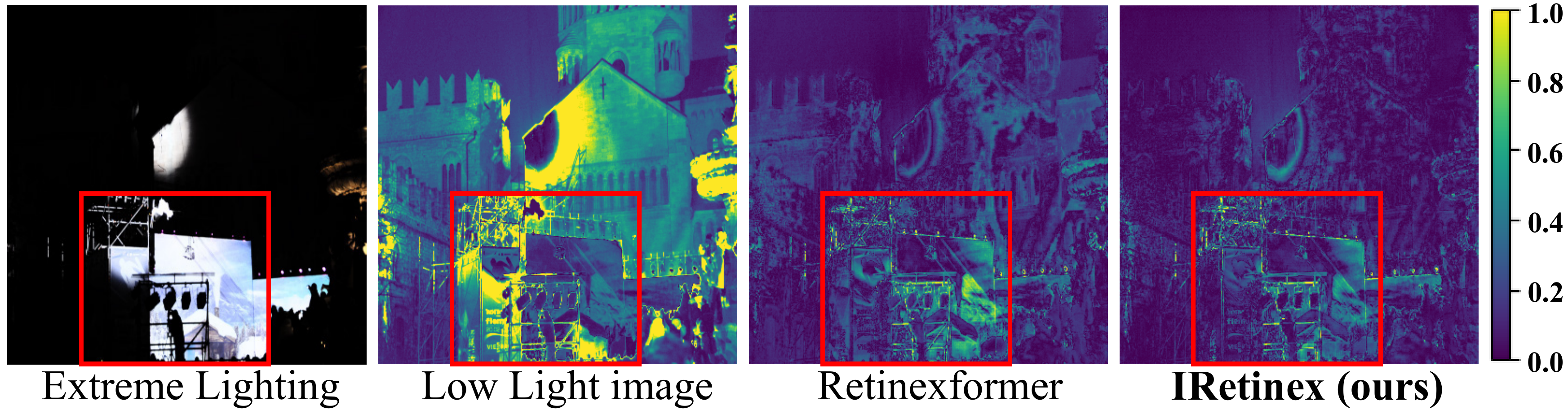}
\caption{The visualization of the structural error maps. It is challenging to simultaneously recover structural information in both underexposed and overexposed regions. }
\label{fig:limitation}
\end{figure}

% \section{Result Credibility}\label{Credibility}
% To ensure the reliability and reproducibility of our findings, we provide comprehensive enhancement results for LOL-v1, LOL-v2-real, and LOL-v2-syn datasets in the accompanying ZIP file. This supplementary material includes the complete set of enhanced images across all benchmark datasets, along with evaluation metric testing code that enables independent verification of our reported performance metrics. The inclusion of both the processed outputs and evaluation framework facilitates transparent assessment of our method's effectiveness and allows for direct comparison with existing approaches under identical testing conditions. 
\begin{figure*}[htpb]
\centering
\includegraphics[width=0.98\textwidth]{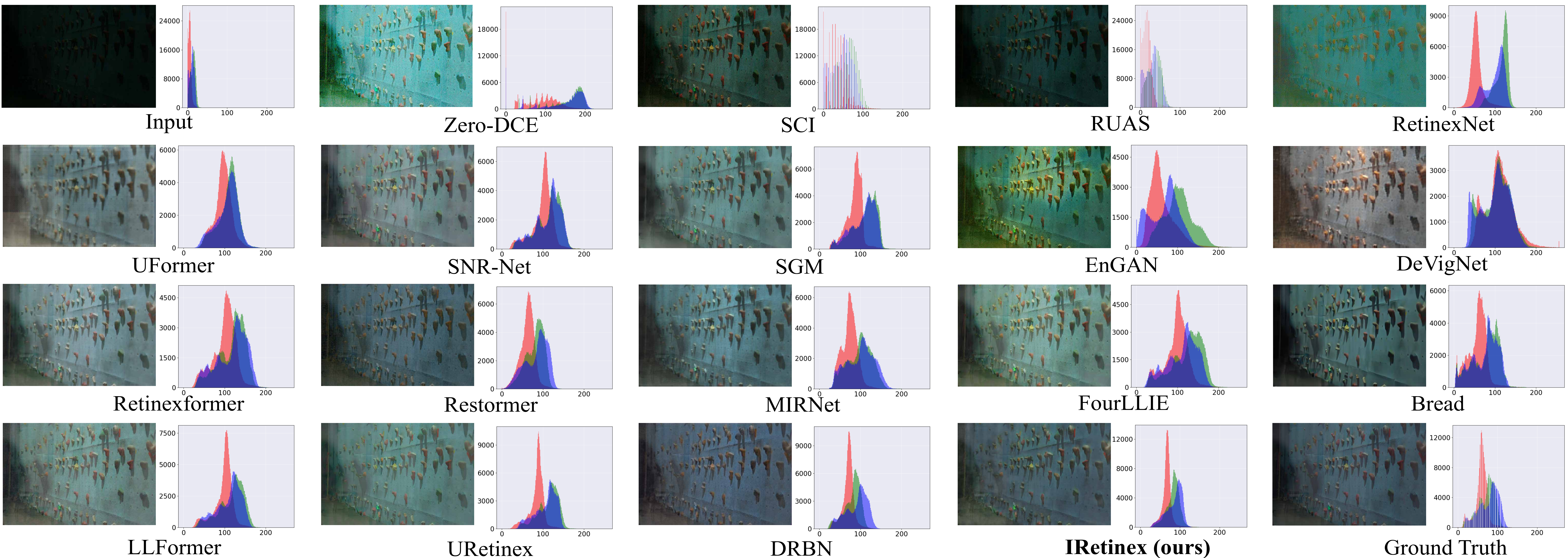} % Reduce the figure size so that it is slightly narrower than the column. Don't use precise values for figure width.This setup will avoid overfull boxes.
\caption{RGB histogram comparison of IRetinex and other LLIE methods on the LOL-v2-real dataset. Red, green, and blue represent R, G, and B color distributions respectively. The x-axis denotes pixel values [0-255], while the y-axis indicates pixel count.}
\label{fig:rgb_more}
\end{figure*}
\begin{figure*}[htpb]
\centering
\includegraphics[width=0.98\textwidth]{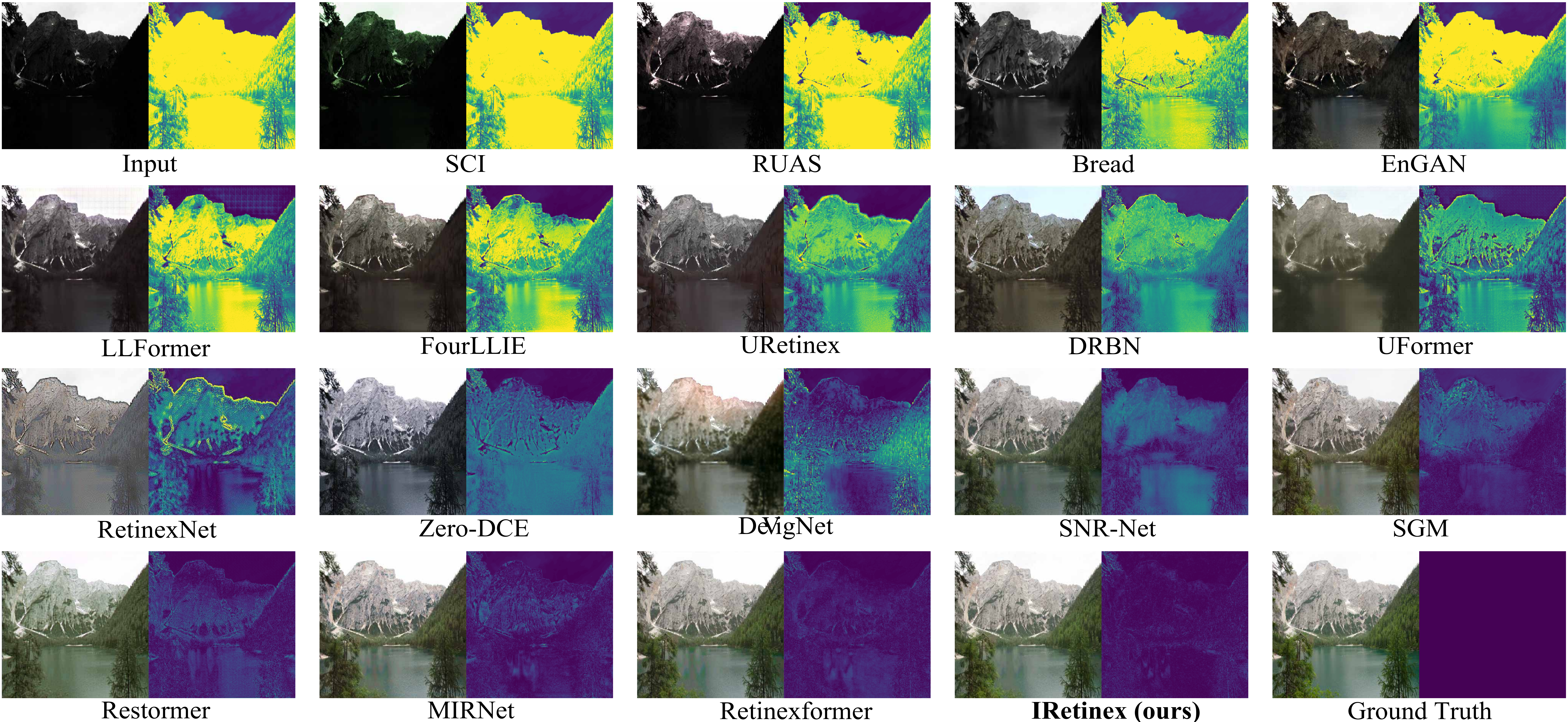} % Reduce the figure size so that it is slightly narrower than the column. Don't use precise values for figure width.This setup will avoid overfull boxes.
\caption{Texture error map comparison between IRetinex and other LLIE methods on the LOL-v2-syn dataset. Color proximity to yellow indicates larger discrepancies, while proximity to blue denotes smaller differences.}
\label{fig:error_more}
\end{figure*}
\section{Discussion and Limitation}\label{Limitation}
Despite IRetinex demonstrating solid performance in both quantitative and qualitative evaluations, it also faces challenges, particularly in handling extreme over-expoursher conditions. For instance, when confronting scenes involving extensive overexposure and extremely low illumination (as illustrated in Fig.~\ref{fig:limitation}), both overexposure and underexposure destroy the original image information. Given the complexity of processing such large illumination differences, our method exhibits notable structural error in overexposed regions. To provide a comparative perspective, we present results obtained from Retinexformer~\cite{caiRetinexformerOnestageRetinexbased} and our methods. Evidently, this limitation is not unique to our method but is inherent to all existing approaches in the LLIE field, representing a shared challenge that requires careful consideration. This challenge stems from the inherent complexity of processing extreme lighting in dynamic scenes and remains an active area of research within the LLIE field. In the future, we will explore the removal of inter-component residuals under extreme lighting conditions.

%%
%% If your work has an appendix, this is the place to put it.

\end{document}